\documentclass[]{ustc_conference}
\usepackage{amssymb}
\usepackage[utf8]{inputenc}
\usepackage{booktabs}
\usepackage{multirow}
\usepackage{xcolor}
\usepackage{colortbl}
\usepackage{adjustbox}
\usepackage{wrapfig}
\usepackage{caption}
\usepackage{needspace}
\usepackage[edges]{forest}
\usepackage{tikz}
\usepackage{amsfonts}
\usepackage[toc,page,header]{appendix}
\usepackage{bbm}
\usepackage{enumitem}
\usepackage{mathtools}
\usepackage[textsize=tiny]{todonotes}
\usepackage[most]{tcolorbox}
\usepackage{makecell}

\definecolor{frameorange}{RGB}{218, 114, 27}
\definecolor{bgorange}{RGB}{253, 243, 235}

\definecolor{frameblue}{RGB}{0, 85, 150}
\definecolor{bgblue}{RGB}{235, 245, 255}

\definecolor{framegreen}{RGB}{34, 139, 34}
\definecolor{bggreen}{RGB}{240, 253, 240}

\newtcolorbox{StrategyBox}[3][frameorange]{
  enhanced,
  float*,
  width=\textwidth,
  title={#3},
  colframe=#1,
  colback=#2,
  colbacktitle=#1,
  coltitle=white,
  fonttitle=\bfseries\large,
  fontupper=\rmfamily,
  arc=1.5mm,
  boxrule=1.2pt,
  top=3mm, bottom=3mm, left=3mm, right=3mm,
  toptitle=0.5mm, bottomtitle=0.5mm,
  before upper={\setlength{\parindent}{1.5em}}
}

\newtcolorbox{BreakableStrategyBox}[3][frameorange]{
  enhanced,
  breakable,
  width=\linewidth,
  title={#3},
  colframe=#1,
  colback=#2,
  colbacktitle=#1,
  coltitle=white,
  fonttitle=\bfseries\large,
  fontupper=\rmfamily,
  arc=1.5mm,
  boxrule=1.2pt,
  top=3mm, bottom=3mm, left=3mm, right=3mm,
  toptitle=0.5mm, bottomtitle=0.5mm,
  before upper={\setlength{\parindent}{1.5em}}
}

\newcounter{idx}

\renewcommand{\paragraph}[1]{\vspace{0.1em}\noindent\textbf{#1}}

\title{CAPO: Critic-Guided Action-Aligned Policy Optimization \\ for Advancing LLM Agent Capabilities}

\author{%
\parbox{\textwidth}{\centering
Daoyu Wang\frontsup{1}, Qingchuan Li\frontsup{1}, Mingyue Cheng\frontsup{1\corrauthor}, Shuo Yu\frontsup{1}, Jie Ouyang\frontsup{1}, \\[2mm]
Chunli Liu\frontsup{2}, Shijin Wang\frontsup{3}, Qi Liu\frontsup{1}, Enhong Chen\frontsup{1}
}}

\affiliation{%
\parbox{\textwidth}
{\centering
\affilsup{1}State Key Laboratory of Cognitive Intelligence, University of Science and Technology of China
}
{\centering
\affilsup{2}School of Management, Hefei University of Technology
}
{\centering
\affilsup{3}iFLYTEK Research
}
}

\date{July 29, 2026}
\correspondence{\email{mycheng@ustc.edu.cn}}

\abstract{
Reinforcement learning (RL) has become a key technique for improving the agentic capabilities of large language models (LLMs). Although critic-free methods such as GRPO are increasingly popular, we argue that critic-based methods remain well suited to long-horizon agentic tasks because their critic models can assess each state and assign credit to different decisions. However, representative critic-based methods such as PPO still organize value estimation, credit assignment, and policy updates at token granularity, whereas the environment state changes only after the agent completes a full action. In this work, we propose CAPO, a critic-guided action-aligned policy optimization method for advancing LLM agent capabilities. CAPO introduces two action-aligned designs. First, it estimates the state value at the action boundary using a critic model and assigns the corresponding advantage to the complete action rather than individual tokens. Second, CAPO introduces a novel action-aware policy ratio that aggregates token ratios within each action with length calibration for consistent updates across actions of different lengths. These two designs effectively achieve fine-grained credit assignment and consistent policy updates while remaining compatible with token-level gradient computation. Experiments across multi-hop QA, academic paper search, and text-world action tasks show that CAPO consistently outperforms representative RL methods. Further analyses validate both designs and reveal how action-aligned optimization improves multi-turn RL training. We hope this work provides a practical path for training more capable LLM agents.
}

\checkdata[Code]{\url{https://github.com/AgentR1/CAPO}}

\begin{document}

\maketitle

\section{Introduction}

Large language models (LLMs) are evolving into autonomous agents capable of interacting with external environments over multiple turns to complete complex tasks \cite{achiam2023gpt,yao2022react,liu2025agentic,cheng2026comprehensive,wang2025paperarena,jiang2026tablemind}. One of the key techniques underlying this evolution is reinforcement learning (RL) \cite{guo2025deepseek,dong2026agentic}, particularly the recent multi-turn RL approaches \cite{zhang2025landscape,wang2026information,tao2026cast,luo2025time}. By treating the LLM as a policy model, multi-turn RL optimizes it over trajectories of long-horizon LLM--environment interactions \cite{qian2026toolrl,cheng2025agentrone}, thereby improving its agentic capabilities to reason over the current observation \cite{wang2025ragen}, make next-step decisions \cite{jin2025search}, and iteratively complete complex tasks \cite{luo2025large}.

\begin{figure}[t]
\centering
\includegraphics[width=\columnwidth]{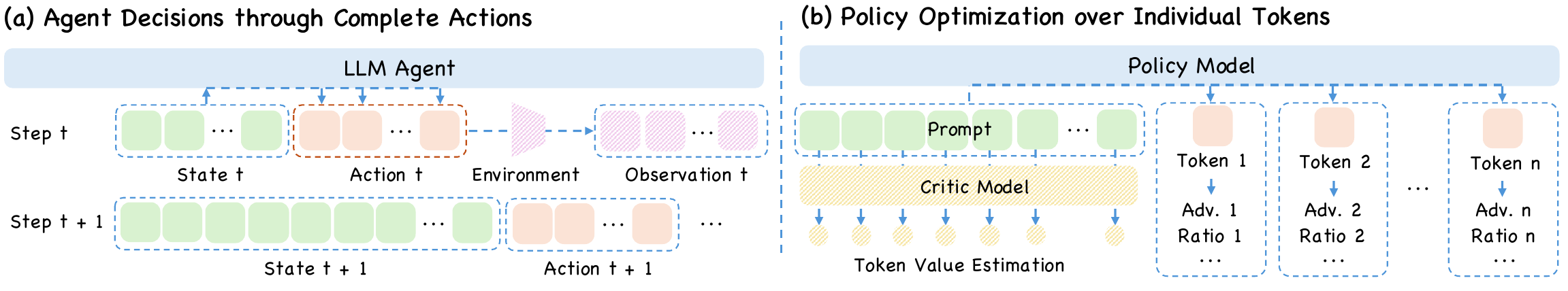}
\caption{LLM agents make decisions through complete actions, but policy optimization is still performed at individual tokens, creating a fundamental granularity mismatch.
}
\label{fig:teaser}
\end{figure}

Recent multi-turn RL methods largely build upon conventional single-turn RL algorithms \cite{song2025r1searcher,xi2025agentgymrl}. Among them, critic-free methods such as GRPO \cite{shao2024deepseekmath} have become increasingly popular because of their simplicity and scalability \cite{li2026step}. These methods avoid training a separate critic model and instead construct relative advantages by comparing groups of sampled trajectories \cite{lin2025cppo}. However, directly extending them to multi-turn training introduces a fundamental limitation: all tokens in a trajectory share one advantage, uniformly reinforcing or penalizing every decision \cite{he2026hierarchy,li2026salt}. To move beyond this coarse trajectory-level signal, recent studies construct advantages from groups of actions under similar states or exploit tree- and graph-structured rollouts for localized comparisons \cite{feng2026gigpo,ji2025treegrpo}. While these approaches provide finer credit signals to some extent, they rely on additional state grouping or structured rollouts, reducing either simplicity or scalability that motivate critic-free methods. Instead, we argue that critic-based methods remain well suited to long-horizon agent tasks because critic models directly estimate state values and enable decision-specific credit signals without modifying rollout trajectories \cite{mnih2016asynchronous,zhou2024archer}. This makes critic-based methods an increasingly attractive alternative for multi-turn LLM agent training \cite{zhang2026reasoning,hou2026single}.

However, as shown in Figure~\ref{fig:teaser}, representative critic-based methods such as PPO \cite{schulman2017ppo} still organize value estimation, credit assignment, and policy updates at token granularity. This creates a granularity mismatch because LLM agents interact with environments through complete actions, each defined as the full response generated before the environment transitions to the next state. Recent studies explore this through step-level formulations, reward designs, or value estimation \cite{li2025st,wei2025reinforcing,li2026turn,wang2026claw}, but mainly address individual components rather than jointly aligning all components at the action granularity. Therefore, an open question arises: \textit{how can critic-based policy optimization be redesigned to better advance LLM agent capabilities by aligning with their decision granularity while remaining computable with token-level RL training?}

In this work, we propose CAPO, a critic-guided action-aligned policy optimization method, where each complete agent action is formulated as the decision unit. CAPO introduces two key designs to perform policy optimization at action granularity: action-level credit assignment and action-aware policy updates. The former leverages a critic model to estimate state values at the action boundary and assigns the corresponding advantages to complete actions rather than individual tokens. The latter introduces a novel action-aware policy ratio that aggregates token ratios with length calibration, enabling consistent weighting and clipping across actions of different lengths. Together, the two designs form an action-aligned optimization objective while remaining compatible with token-level gradient computation.

The key insight behind CAPO is that LLM agents make decisions through complete actions, whereas tokens only serve as the underlying generation units. Accordingly, all components of policy optimization should be aligned with action granularity while remaining compatible with token-level gradient computation. This insight motivates two designs above. First, credit should reflect the contribution of each complete action rather than be spread across individual tokens or shared over an entire trajectory. Second, policy updates should weight and clip each complete action rather than each token within it. Additionally, because actions vary in length, directly aggregating token ratios through multiplication or averaging can introduce length-dependent inconsistency in weighting and clipping, motivating length calibration for more comparable and consistent updates across different actions.

We evaluate CAPO across multi-hop question answering (QA), academic paper search, and text-world action tasks. Experimental results show that CAPO consistently outperforms representative RL algorithms with different decision granularities, credit assignment strategies, and policy-update designs. Further analyses, including ablation and sensitivity studies, validate the effectiveness of both action-level credit assignment and action-aware policy updates, providing insights into how action-aligned optimization improves multi-turn RL training. We hope CAPO offers a useful and practical path toward training more capable LLM agents.

In summary, our contributions are as follows:

\begin{itemize}
    \item We revisit the underexplored potential of critic-based methods for multi-turn LLM agents despite the popularity of critic-free RL methods, and identify the granularity mismatch that limits this potential.
    \item We propose CAPO, a critic-based policy optimization method that aligns credit assignment and policy updates with complete agent actions while remaining compatible with token-level gradient computation.
    \item We conduct extensive experiments across diverse agent tasks, showing consistent improvements and providing insights into action-aligned optimization for LLM agents.
\end{itemize}

\section{Related Work}

In this section, we review advances in reinforcement learning (RL) for LLMs and its extension to multi-turn agent tasks. We focus on how existing methods choose the granularity of credit signals and policy updates, which motivates aligning both with complete agent actions.

\subsection{Reinforcement Learning for LLMs}

Reinforcement Learning has become a major post-training method \cite{ouyang2022training,guo2025deepseek}. Recent studies increasingly favor critic-free methods such as GRPO \cite{shao2024deepseekmath} because of their simple formulation and high scalability \cite{ahmadian2024back,hu2025reinforce++}. These methods avoid training a separate critic and construct relative advantages by comparing sampled responses from the same prompt, enabling efficient optimization for long-form reasoning \cite{yu2026dapo,xu2025toward}. However, they typically produce coarse credit signals that cannot distinguish intermediate decisions in multi-step tasks \cite{liu2026shadow}. Recent extensions derive more localized signals through additional state grouping or structured rollouts, but partly sacrifice the simplicity \cite{ji2025treegrpo,feng2026gigpo}. This limitation motivates renewed attention to critic-based methods such as PPO \cite{schulman2017ppo}, whose critic models directly estimate state values and can provide decision-specific credit signals for long-horizon tasks, despite additional GPU memory overhead \cite{hou2026single}. Critic-based and critic-free methods therefore offer complementary trade-offs: the former can localize credit through value estimates, while the latter avoids critic training but relies more heavily on the organization of rollout groups.

\subsection{Multi-Turn Reinforcement Learning}

Multi-turn RL largely extends conventional single-turn RL algorithms by optimizing over trajectories of long-horizon LLM--environment interactions \cite{wang2025ragen,zhang2026instinct,guo2026segment,pan2026paperscout}. This transition motivates new designs in credit assignment to handle sparse rewards and in policy updates to stabilize long-horizon optimization \cite{ji2025treegrpo,zong2026at2po,li2026ssvpo}. For credit assignment, recent studies such as GiGPO \cite{feng2026gigpo}, ST-PPO \cite{li2025st}, and Turn-PPO \cite{li2026turn} begin to explore finer-grained credit assignment in multi-turn settings and show the benefits of more localized feedback. For policy updates, GSPO \cite{zheng2025group} and GMPO \cite{zhao2026geometric} replace token-wise policy ratios with sequence-level ratios to reduce noisy token weighting and stabilize clipping. CTPO \cite{zhang2026rethinking} further studies the trade-off between local token correction and complete-sequence importance sampling. These lines of work improve either the signal assigned to a decision or the ratio used to update it, but do not generally make both choices at the same interaction granularity. In contrast, CAPO jointly aligns credit assignment and policy updates with the granularity of complete agent actions, providing a unified action-aligned optimization.

\begin{figure*}[t]
\centering
\includegraphics[width=\textwidth]{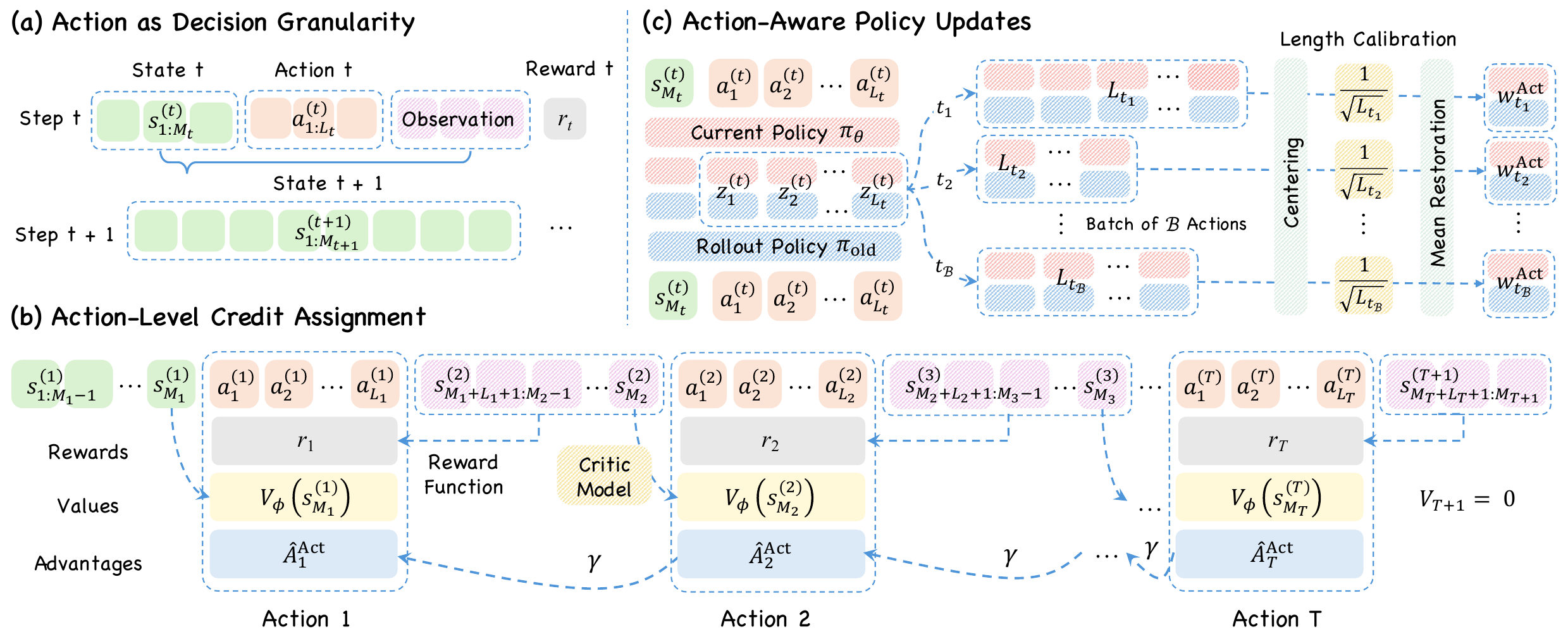}
\caption{Overview of CAPO. CAPO treats each complete action as the decision granularity, assigns action-level credit along interaction steps, and performs action-aware policy updates through a length-calibrated policy ratio across action lengths.}
\label{fig:framework}
\end{figure*}

\section{Methodology}

In this section, we present an overview of CAPO, formulate the decision granularity, decompose its two key designs and show the action-aligned optimization objective. The construction first assigns credit to complete actions and then uses an action-level ratio for weighting and clipping, while retaining the token score gradients required by autoregressive policy optimization.

\subsection{Overview}
The motivation of CAPO is that critic-based methods provide state-aware credit signals for multi-turn RL, but their potential is limited by the mismatch between agent decisions and token-level optimization. As shown in Figure~\ref{fig:framework}, CAPO treats complete actions as the agent decision granularity and introduces two designs. The first uses the critic model to estimate state values and assign advantages to complete actions. The second constructs an action-aware policy ratio for length-calibrated weighting and clipping of complete actions. The two designs address distinct stages of the actor update: the advantage specifies which action is reinforced, whereas the ratio specifies how that action is weighted and clipped. Together, these designs form an action-aligned objective while remaining compatible with token-level gradient computation.

\subsection{Action as Decision Granularity}

As illustrated in Figure \ref{fig:framework}(a), at interaction step $t$, the state $s^{(t)}_{1:M_t}$ denotes the interaction history available to the agent, including previous observations and actions, with $M_t$ tokens. The action $a^{(t)}_{1:L_t}$ is a complete response generated by the policy, containing $L_t$ reasoning and tool-call tokens in common. The policy interacts with the environment over a horizon of $T$ steps. At each step $t$, the policy observes $s^{(t)}_{1:M_t}$, emits $a^{(t)}_{1:L_t}$, receives reward $r_t$, and then transitions to the next state $s^{(t+1)}_{1:M_{t+1}}$. These steps form the trajectory $\tau$:
\begin{equation}
\begin{aligned}
\tau
&=\{(s^{(t)}_{1:M_t},a^{(t)}_{1:L_t},r_t,s^{(t+1)}_{1:M_{t+1}})\}_{t=1}^{T},
\end{aligned}
\label{eq:step_mdp}
\end{equation}
where newly generated tokens are appended to the action prefix autoregressively until termination. The environment transition occurs only after the complete action, and subsequent observations are appended to the next state.

\subsection{Action-Level Credit Assignment}

CAPO computes credit at the action granularity of agent decisions by assigning an advantage to each complete action. Specifically, the advantage compares the expected return after taking the action with the value of the pre-action state. Therefore, CAPO estimates $V_\phi(s^{(t)}_{1:M_t})$ before generating $a^{(t)}_{1:L_t}$ using the critic model. Notably, value estimation at this action boundary requires no additional state encoder. Since the final token of $s^{(t)}_{1:M_t}$ is conditioned on the entire state prefix and lies at the action boundary, CAPO uses its value prediction as the state value, i.e., $V_\phi(s^{(t)}_{1:M_t}) \triangleq V_\phi(s^{(t)}_{M_t})$, where $V_\phi$ denotes the critic parameterized by $\phi$. Furthermore, $r_t$ is defined as the reward of this action, which may consist of intermediate environment feedback or summed token penalties. Even with terminal-only rewards, the critic produces distinct advantages through bootstrapped value differences. As shown in Figure~\ref{fig:framework}(b), CAPO orders the steps within each trajectory and applies Generalized Advantage Estimation (GAE) \cite{schulman2015high} along the resulting timeline to compute the action-level advantage $\hat{A}_t^{\mathrm{Act}}$ as follows:
\begin{equation}
\begin{aligned}
\hat{A}_t^{\mathrm{Act}}
&= \sum_{l=0}^{T-t}
(\gamma\lambda)^l\delta_{t+l}, \\
\delta_t
&= r_t+\gamma V_\phi(s^{(t+1)}_{M_{t+1}})
-V_\phi(s^{(t)}_{M_t}),
\end{aligned}
\label{eq:action_gae}
\end{equation}
where $\gamma$ is the discount factor, $\delta_t$ is the temporal-difference (TD) residual and $\lambda$ denotes the trace parameter. For the terminal step, the value of the subsequent state is set to zero. The advantage $\hat{A}_t^{\mathrm{Act}}$ is assigned to the complete action and shared by its action tokens for the actor update. Compared with token-level credit that may not reflect complete-action effects and trajectory-level credit that shares one signal across the trajectory, action-level credit provides more informative signals for intermediate decisions in long-horizon interactions.

\subsection{Action-Aware Policy Updates}

Action-level credit assignment provides one advantage for each complete agent action. However, PPO- and GRPO-style objectives still update the policy using token-wise weighting and clipping. Specifically, the importance sampling ratio for the $i$-th token in the action at interaction step $t$ is defined as:
\begin{equation}
w_i^{(t)}(\theta)
=
\frac{
\pi_\theta(a^{(t)}_i\mid s^{(t)}_{1:M_t},a^{(t)}_{<i})
}{
\pi_{\mathrm{old}}(a^{(t)}_i\mid s^{(t)}_{1:M_t},a^{(t)}_{<i})
},
\label{eq:token_ratio}
\end{equation}
where $\pi_\theta$ and $\pi_{\mathrm{old}}$ denote the current and rollout policies, respectively. For convenience, we define the log-ratio as
$z_i^{(t)}(\theta)\triangleq\log w_i^{(t)}(\theta)$.
Although all tokens within an action share the action-level advantage, token-wise weighting and clipping still operate on individual tokens instead of complete actions \cite{lei2026step}.

\begin{wrapfigure}{r}{0.52\linewidth}
\centering
\includegraphics[width=\linewidth]{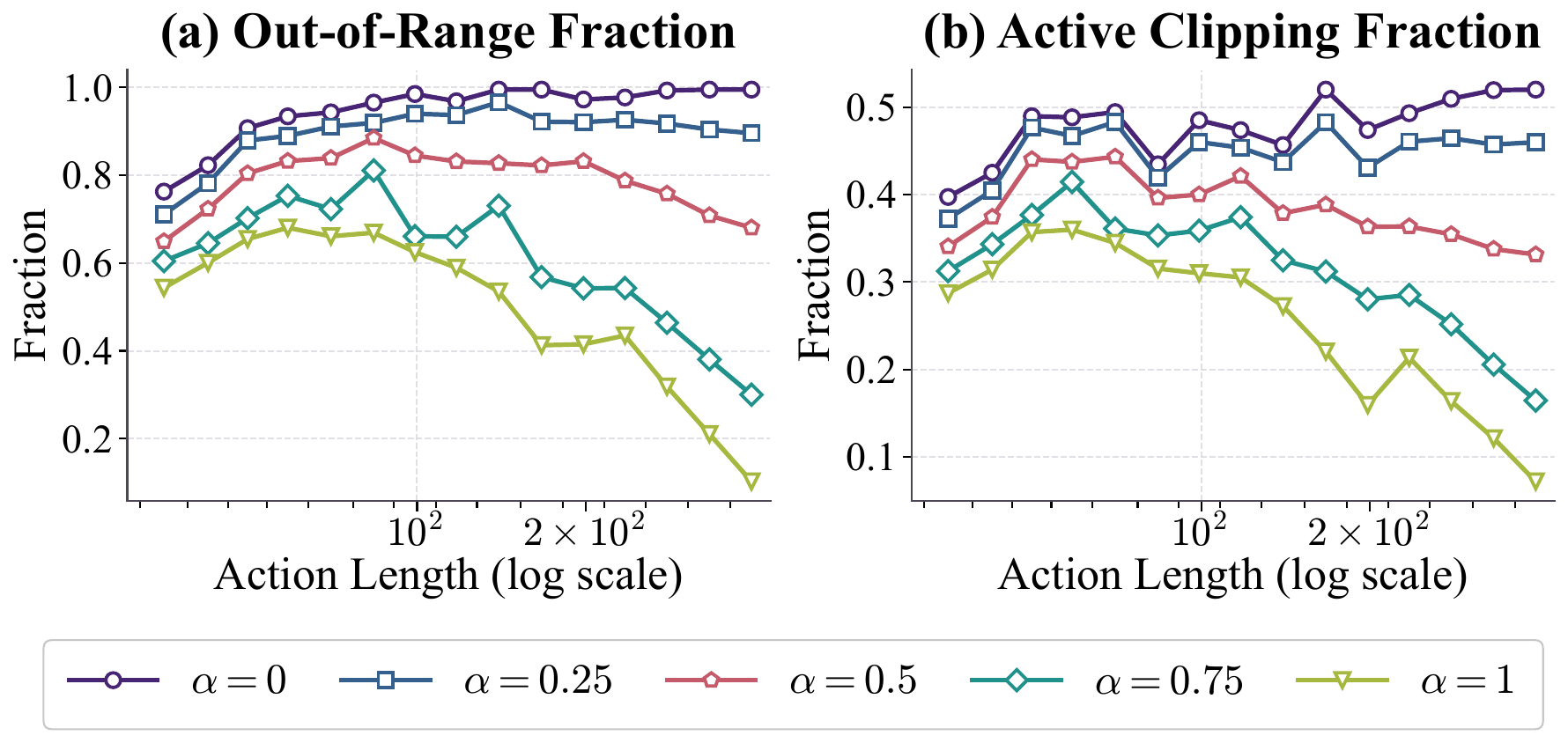}
\caption{Toy experiment with different $\alpha$ across action lengths. (a) Out-of-range fraction: the proportion of length-scaled ratios that fall outside $[1-\epsilon,1+\epsilon]$. (b) Active clipping fraction: the proportion for which clipping is activated in the policy objective. We set $\epsilon=0.001$ in this experiment.
}
\label{fig:clipping}
\end{wrapfigure}

The most straightforward way to construct an action-level policy ratio is to multiply the token-wise ratios of the complete action, yielding the exact action likelihood ratio
$\prod_{i=1}^{L_t}w_i^{(t)}(\theta)
=\exp(\sum_{i=1}^{L_t}z_i^{(t)}(\theta))$.
However, its scale becomes increasingly sensitive to action length \cite{zhang2026rethinking}. Existing aggregation methods, such as the geometric mean
$\exp(\frac{1}{L_t}\sum_{i=1}^{L_t}z_i^{(t)}(\theta))$,
effectively control this growth but can result in substantially lower clipping fractions for longer actions \cite{zheng2025group}.
More generally, the two approaches above can be unified within the length-scaled family
$\exp(\frac{1}{L_t^\alpha}\sum_{i=1}^{L_t}z_i^{(t)}(\theta))$,
where $\alpha=0$ corresponds to the former and $\alpha=1$ to the latter. In Figure~\ref{fig:clipping}, we compare $\alpha\in\{0,0.25,0.5,0.75,1\}$ by plotting the out-of-range fraction and active clipping fraction across action lengths. Consistent with the discussion above, both fractions increase with length for $\alpha=0$ and decrease for $\alpha=1$, indicating length-dependent clipping bias, whereas $\alpha=0.5$ yields more comparable fractions across action lengths.

Motivated by this observation, CAPO introduces a novel action-aware policy ratio that calibrates the update scale across action lengths using square-root scaling over centered token-level log-ratios. As shown in Figure~\ref{fig:framework}(c), let $\mathcal{B}$ denote the set of actions in the current optimization batch, and let $\operatorname{sg}(\cdot)$ denote the stop-gradient operation. We define:
\begin{equation}
\begin{aligned}
w_t^{\mathrm{Act}}(\theta)
&=
\exp\Bigg(
\hat{\mu}(\theta)
+
\frac{
\sum_{i=1}^{L_t}
\bigl(z_i^{(t)}(\theta)-\hat{\mu}(\theta)\bigr)
}{
\sqrt{L_t}
}
\Bigg),\\
\hat{\mu}(\theta)
&=
\operatorname{sg}\left(
\frac{
\sum_{b\in\mathcal{B}}\sum_{i=1}^{L_b}z_i^{(b)}(\theta)
}{
\sum_{b\in\mathcal{B}}L_b
}
\right),
\end{aligned}
\label{eq:action_aware_ratio}
\end{equation}
where $w_t^{\mathrm{Act}}(\theta)$ is a length-calibrated policy ratio rather than an exact action likelihood ratio. Under an i.i.d.\ approximation in which the centered token-level log-ratios have variance $\sigma^2$, the normalized sum has variance $\frac{1}{L_t}\cdot L_t\sigma^2=\sigma^2$, independent of action length. This provides the intuition for choosing $\alpha=1/2$ in the proposed length calibration. The centering term further removes the length-dependent mean shift introduced by direct square-root aggregation: if $\mathbb{E}[z_i^{(t)}(\theta)]=\mu$, direct aggregation has mean $\sqrt{L_t}\mu$, whereas centering before scaling and restoring the batch mean approximately preserves the original center. The resulting ratio therefore yields more consistent clipping fractions and more comparable policy updates across actions of different lengths.

\subsection{Action-Aligned Policy Optimization Objective}

Finally, CAPO combines the action-level advantage $\hat{A}_t^{\mathrm{Act}}$ with the action-aware policy ratio $w_t^{\mathrm{Act}}(\theta)$ to define an action-aligned optimization objective:
\begin{equation}
\begin{aligned}
\mathcal{J}_{\mathrm{CAPO}}(\theta)
&= 
\mathbb{E}_{\tau\sim\pi_{\mathrm{old}}}
\Big[
\min\Big(
w_t^{\mathrm{Act}}(\theta)
\hat{A}_t^{\mathrm{Act}},\\
&\operatorname{clip}\big(
w_t^{\mathrm{Act}}(\theta),
1-\epsilon,1+\epsilon
\big)
\hat{A}_t^{\mathrm{Act}}
\Big)
\Big],
\end{aligned}
\label{eq:capo_objective}
\end{equation}
where $\epsilon$ is the clipping parameter. As $\hat{\mu}(\theta)$ is stop-gradient, it affects the ratio and clipping without introducing additional derivative terms through the shared batch statistic. Treating $\hat{A}_t^{\mathrm{Act}}$ as fixed during the actor update, before clipping is activated the gradient is:
\begin{equation}
\begin{aligned}
\nabla_\theta\!\left(
w_t^{\mathrm{Act}}(\theta)\hat{A}_t^{\mathrm{Act}}
\right)
&=
\frac{w_t^{\mathrm{Act}}(\theta)\hat{A}_t^{\mathrm{Act}}}{\sqrt{L_t}}
\sum_{i=1}^{L_t}\nabla_\theta z_i^{(t)}(\theta)\\
&=
\frac{w_t^{\mathrm{Act}}(\theta)\hat{A}_t^{\mathrm{Act}}}{\sqrt{L_t}}
\sum_{i=1}^{L_t}
\nabla_\theta\log\pi_\theta\!\left(
a_i^{(t)}\mid s^{(t)}_{1:M_t},a_{<i}^{(t)}
\right),
\end{aligned}
\label{eq:capo_gradient}
\end{equation}
where the second equality follows because the rollout policy is fixed. Thus, token score gradients within an action are summed under one action advantage, action ratio, and clipping decision; $L_t^{-1/2}$ calibrates their magnitude across action lengths. When $L_t=1$, the action-aware ratio reduces to the standard token ratio. Near the rollout policy, CAPO therefore follows the summed token-score direction for the action while preventing longer actions from dominating solely through token count. Because $w_t^{\mathrm{Act}}(\theta)$ is a length-calibrated surrogate rather than the exact complete-action likelihood ratio, this analysis describes its gradient weighting and clipping behavior rather than an unbiased importance-sampling correction. The action-level advantage captures the contribution of each action, while action-aware policy updates provide length-calibrated weighting and clipping across actions with different lengths. Both designs operate at the action level while remaining compatible with token-level gradient computation.

\section{Experiments}
\label{sec:experiments}

In this section, we present the overall performance of CAPO and conduct in-depth analyses of its credit assignment and policy-update designs. We evaluate both task outcomes and optimization behavior under shared task interfaces, training data, and evaluation protocols.

\begin{table*}[t]
    \centering
    \caption{Main results on multi-hop QA, academic paper search, ALFWorld, and WebShop, reported as mean$_{\pm\mathrm{std}}$. HotpotQA is in-domain ($\dagger$), while 2Wiki and MuSiQue are out-of-domain ($\star$). Bold and underlining denote the best and second-best results.}
    \label{tab:main_results}
    \resizebox{\textwidth}{!}{
    \begin{tabular}{l|ccc|cc|cc|cc}
    \toprule
    
    \multirow{2}{*}{\textbf{Method}}
    & \multirow{2}{*}{\textbf{HotpotQA}$^{\dagger}$}
    & \multirow{2}{*}{\textbf{2Wiki}$^{\star}$}
    & \multirow{2}{*}{\textbf{MuSiQue}$^{\star}$}
    & \multicolumn{2}{c|}{\textbf{RealResearchQuery}}
    & \multicolumn{2}{c|}{\textbf{ALFWorld}}
    & \multicolumn{2}{c}{\textbf{WebShop}} \\
    
    \cmidrule(lr){5-6}
    \cmidrule(lr){7-8}
    \cmidrule(lr){9-10}
    
    & & &
    & \textbf{F1@all}
    & \textbf{Recall@all}
    & \textbf{Seen}
    & \textbf{Unseen}
    & \textbf{Score}
    & \textbf{Succ.} \\
    
    \midrule
    
    \multicolumn{10}{l}{\textit{Qwen3-1.7B}} \\
    
    \hspace{0.5em}+ ReAct
    & 3.62 & 2.24 & 0.50
    & 0.005 & 0.012
    & 2.86 & 2.24
    & 2.03 & 0.80 \\
    
    \hspace{0.5em}+ PPO
    & 38.00$_{\pm0.75}$ & 50.12$_{\pm1.02}$ & 16.55$_{\pm0.83}$
    & 0.284$_{\pm0.004}$ & 0.514$_{\pm0.004}$
    & 67.14$_{\pm0.60}$ & 69.40$_{\pm2.14}$
    & 59.12$_{\pm0.78}$ & 34.60$_{\pm1.34}$ \\
    
    \hspace{0.5em}+ Reinforce++
    & 37.92$_{\pm2.62}$ & 49.01$_{\pm0.67}$ & 17.09$_{\pm0.31}$
    & 0.281$_{\pm0.001}$ & 0.521$_{\pm0.006}$
    & 65.00$_{\pm0.03}$ & 68.66$_{\pm2.00}$
    & 61.34$_{\pm2.18}$ & 35.40$_{\pm0.88}$ \\
    
    \hspace{0.5em}+ GRPO
    & 36.76$_{\pm1.25}$ & 48.30$_{\pm1.35}$ & 16.88$_{\pm0.76}$
    & 0.275$_{\pm0.003}$ & 0.530$_{\pm0.007}$
    & \underline{73.57}$_{\pm0.06}$ & \underline{75.37}$_{\pm2.24}$
    & 63.15$_{\pm1.85}$ & 36.20$_{\pm0.60}$ \\
    
    \hspace{0.5em}+ RLOO
    & 37.41$_{\pm0.83}$ & 47.26$_{\pm0.23}$ & 15.93$_{\pm0.77}$
    & 0.279$_{\pm0.001}$ & 0.536$_{\pm0.005}$
    & 71.43$_{\pm1.59}$ & 73.88$_{\pm0.76}$
    & 65.74$_{\pm0.15}$ & 35.00$_{\pm0.83}$ \\
    
    \hspace{0.5em}+ GSPO
    & 39.12$_{\pm2.33}$ & 46.83$_{\pm0.68}$ & 17.46$_{\pm2.77}$
    & 0.264$_{\pm0.008}$ & 0.541$_{\pm0.002}$
    & 67.86$_{\pm0.27}$ & 67.91$_{\pm1.22}$
    & 59.03$_{\pm1.47}$ & 32.40$_{\pm1.05}$ \\
    
    \hspace{0.5em}+ GiGPO
    & 40.85$_{\pm2.79}$ & \underline{52.43}$_{\pm0.67}$ & 18.37$_{\pm0.26}$
    & \underline{0.298}$_{\pm0.010}$ & 0.545$_{\pm0.002}$
    & 70.00$_{\pm1.24}$ & 69.40$_{\pm1.11}$
    & \underline{66.92}$_{\pm0.65}$ & \underline{41.80}$_{\pm1.83}$ \\
    
    \hspace{0.5em}+ ST-PPO
    & \underline{41.32}$_{\pm0.22}$ & 51.76$_{\pm1.29}$ & \underline{18.91}$_{\pm0.10}$
    & 0.294$_{\pm0.003}$ & \underline{0.548}$_{\pm0.004}$
    & 69.29$_{\pm0.94}$ & 70.90$_{\pm3.16}$
    & 65.80$_{\pm0.59}$ & 40.60$_{\pm1.11}$ \\
    
    \rowcolor{gray!15}
    \hspace{0.5em}+ \textbf{CAPO}
    & \textbf{45.19}$_{\pm0.54}$ & \textbf{56.64}$_{\pm1.23}$ & \textbf{21.89}$_{\pm0.70}$
    & \textbf{0.319}$_{\pm0.007}$ & \textbf{0.557}$_{\pm0.002}$
    & \textbf{75.71}$_{\pm0.59}$ & \textbf{79.85}$_{\pm0.55}$
    & \textbf{70.40}$_{\pm0.70}$ & \textbf{45.60}$_{\pm0.58}$ \\
    
    \midrule
    
    \multicolumn{10}{l}{\textit{Qwen3-4B}} \\
    
    \hspace{0.5em}+ ReAct
    & 37.45 & 48.59 & 10.26
    & 0.171 & 0.193
    & 7.14 & 2.99
    & 51.58 & 23.80 \\
    
    \hspace{0.5em}+ PPO
    & 56.75$_{\pm1.02}$ & 58.92$_{\pm1.16}$ & 19.82$_{\pm0.81}$
    & 0.303$_{\pm0.005}$ & 0.531$_{\pm0.002}$
    & 76.43$_{\pm0.09}$ & 72.39$_{\pm1.16}$
    & 70.18$_{\pm2.71}$ & 46.00$_{\pm3.35}$ \\
    
    \hspace{0.5em}+ Reinforce++
    & 55.94$_{\pm0.67}$ & 60.48$_{\pm0.97}$ & 21.72$_{\pm1.01}$
    & 0.286$_{\pm0.003}$ & 0.548$_{\pm0.002}$
    & 72.86$_{\pm1.92}$ & 71.64$_{\pm0.20}$
    & 67.84$_{\pm2.25}$ & 47.20$_{\pm0.71}$ \\
    
    \hspace{0.5em}+ GRPO
    & 56.61$_{\pm1.03}$ & \underline{63.33}$_{\pm0.62}$ & \underline{25.07}$_{\pm0.85}$
    & 0.294$_{\pm0.003}$ & \underline{0.572}$_{\pm0.005}$
    & 81.43$_{\pm2.16}$ & 74.63$_{\pm1.09}$
    & 65.83$_{\pm0.89}$ & 44.20$_{\pm1.43}$ \\
    
    \hspace{0.5em}+ RLOO
    & 56.31$_{\pm1.66}$ & 61.85$_{\pm0.86}$ & 23.91$_{\pm0.81}$
    & 0.297$_{\pm0.003}$ & 0.563$_{\pm0.008}$
    & 75.71$_{\pm2.10}$ & 70.90$_{\pm0.82}$
    & 62.49$_{\pm1.12}$ & 42.60$_{\pm0.71}$ \\
    
    \hspace{0.5em}+ GSPO
    & 57.08$_{\pm1.49}$ & 56.14$_{\pm2.68}$ & 22.59$_{\pm0.13}$
    & 0.289$_{\pm0.007}$ & 0.552$_{\pm0.002}$
    & 79.29$_{\pm1.76}$ & 77.61$_{\pm1.00}$
    & 69.78$_{\pm0.24}$ & 48.80$_{\pm2.09}$ \\
    
    \hspace{0.5em}+ GiGPO
    & 58.14$_{\pm0.78}$ & 61.27$_{\pm0.16}$ & 23.50$_{\pm2.04}$
    & \underline{0.306}$_{\pm0.003}$ & 0.567$_{\pm0.001}$
    & \underline{88.57}$_{\pm2.59}$ & 79.10$_{\pm2.19}$
    & 67.13$_{\pm0.74}$ & \underline{50.00}$_{\pm0.75}$ \\
    
    \hspace{0.5em}+ ST-PPO
    & \underline{59.01}$_{\pm2.41}$ & 60.84$_{\pm2.79}$ & 24.16$_{\pm3.34}$
    & 0.302$_{\pm0.004}$ & 0.571$_{\pm0.002}$
    & 87.86$_{\pm1.75}$ & \underline{80.60}$_{\pm0.80}$
    & \underline{70.60}$_{\pm0.87}$ & 49.60$_{\pm0.98}$ \\
    
    \rowcolor{gray!15}
    \hspace{0.5em}+ \textbf{CAPO}
    & \textbf{64.12}$_{\pm1.22}$ & \textbf{66.48}$_{\pm1.12}$ & \textbf{27.16}$_{\pm0.84}$
    & \textbf{0.318}$_{\pm0.002}$ & \textbf{0.590}$_{\pm0.002}$
    & \textbf{92.86}$_{\pm1.44}$ & \textbf{86.57}$_{\pm1.00}$
    & \textbf{78.12}$_{\pm1.17}$ & \textbf{58.20}$_{\pm0.56}$ \\
    
    \bottomrule
    \end{tabular}
    }
\end{table*}

\subsection{Experimental Setup}
\paragraph{Backbones.}
We evaluate CAPO on two backbone models, Qwen3-1.7B and Qwen3-4B-Instruct-2507 \cite{yang2025qwen3}. Across all tasks, we adopt an OpenAI-compatible tool calling format. The prompt includes the current observation, interaction history, available tools, and output format.

\paragraph{Benchmarks.}
We evaluate CAPO on four agent tasks. For multi-hop QA, we use HotpotQA \cite{yang2018hotpotqa} as the in-domain benchmark and 2Wiki \cite{ho2020constructing} and MuSiQue \cite{trivedi2022musique} for out-of-domain generalization. The agent performs Wikipedia search and generates short answers, and we report answer accuracy (Acc). For academic paper search, we evaluate on RealResearchQuery \cite{he2025pasa} under the same setting of PaperScout \cite{pan2026paperscout}. The agent performs paper search and citation/reference expansion over a maintained paper pool; we report F1@all and Recall@all. For ALFWorld \cite{shridhar2020alfworld}, the agent solves household tasks in text-based embodied environments through textual commands, and we report win rates on seen and unseen validation splits. For WebShop \cite{yao2022webshop}, the agent navigates a text-based shopping environment with actions including search, selection, and purchase; we report average task score (Score) and purchase success rate (Succ.). Across all tasks, rewards combine terminal correctness with intermediate signals for valid tool-call formatting.

\paragraph{Baselines.}
We compare against various baselines. The prompting baselines include ReAct \cite{yao2022react}. RL baselines include PPO \cite{schulman2017ppo}, Reinforce++ \cite{hu2025reinforce++}, GRPO \cite{shao2024deepseekmath}, RLOO \cite{ahmadian2024back}, GSPO \cite{zheng2025group}, GiGPO \cite{feng2026gigpo} and ST-PPO \cite{li2025st}. All RL baselines share the same prompt, training data, and evaluation protocols.

\paragraph{Training Details.}
Our implementation uses Agent-R1 \cite{cheng2025agentrone} as the training framework, with veRL \cite{sheng2025hybridflow} as the backend RL engine. Experiments are run with 8 NVIDIA H100 GPUs, and all RL results are averaged over three random seeds. Unless otherwise specified, we set the actor learning rate to \(1\times10^{-6}\), the critic learning rate to \(1\times10^{-5}\) for critic-based methods, the training batch size to $128$, and the actor micro-batch size to $4$ per GPU. Group-based baselines use $8$ rollouts per prompt and a batch size of $16$, yielding an effective batch size of $128$ for fair comparison. For CAPO, we use discount factor $\gamma = 0.99$, trace parameter $\lambda = 1.0$, KL regularization coefficient to $0.001$, and set the symmetric clipping parameter to $\epsilon = 0.001$.

\subsection{Main Results}
Table~\ref{tab:main_results} summarizes the overall performance across multi-hop QA, academic paper search, and text-world action tasks. CAPO achieves the best results on all metrics for both backbone models, with substantial improvements on long-horizon tasks. On multi-hop QA, CAPO obtains consistent gains on in-domain HotpotQA and out-of-domain 2Wiki and MuSiQue, showing stronger generalization under distribution shifts. The improvements are more pronounced on agent tasks, where effective credit assignment over interaction steps is critical. Notably, CAPO surpasses GiGPO and ST-PPO, which also exploit action-level modeling but retain token-wise or sequence-aggregated policy ratios. These results demonstrate that aligning policy optimization with action granularity is essential for training capable LLM agents.

\begin{figure}[t]
\centering
\includegraphics[width=\linewidth]{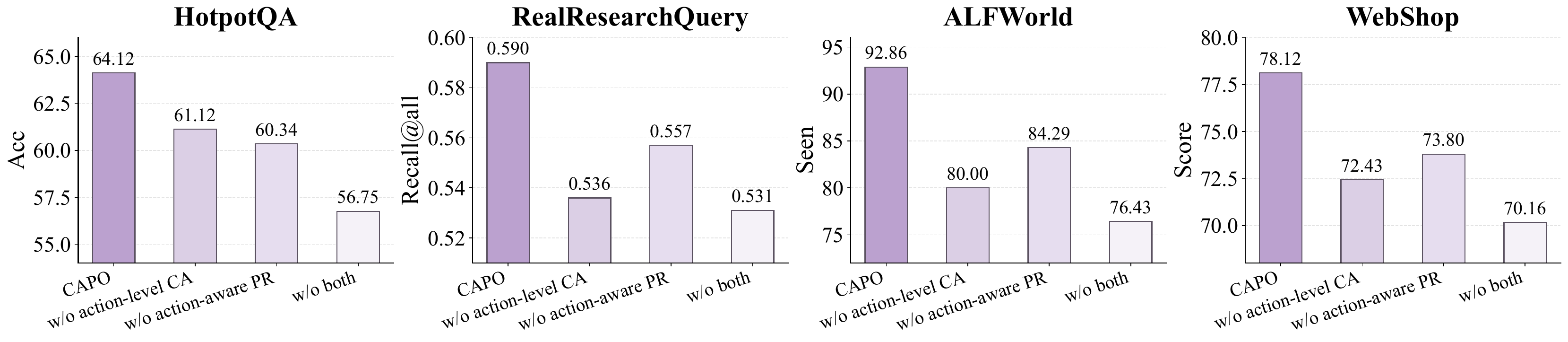}
\caption{Ablation across four benchmarks. Removing action-level credit assignment (CA) or action-aware policy updates (PU) hurts performance, particularly without both.}
\label{fig:ablation}
\end{figure}

\begin{figure*}[t]
\centering
\includegraphics[width=\linewidth]{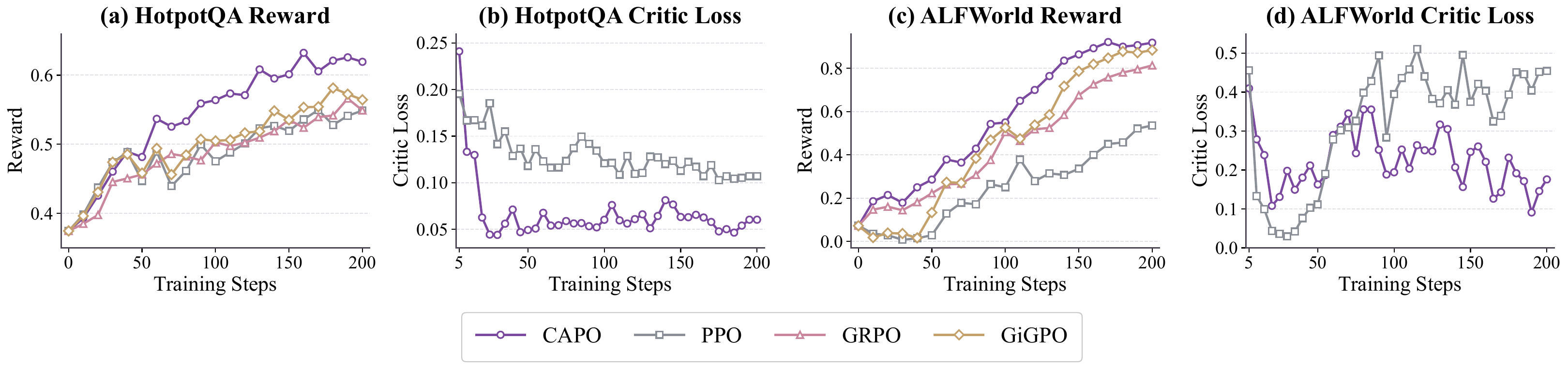}
\caption{
Training dynamics on HotpotQA and ALFWorld. CAPO achieves higher rewards than PPO, GRPO, and GiGPO while maintaining lower critic loss than PPO after the initial updates, indicating more efficient learning and more stable optimization.
}
\label{fig:training_curves}
\end{figure*}

\subsection{Ablation Study}
We ablate CAPO's action-level credit assignment and action-aware policy updates. The former replaces interaction-step GAE with PPO token-level GAE; the latter restores token-wise ratios and clipping. Figure~\ref{fig:ablation} shows that credit assignment contributes more on the interaction-intensive RealResearchQuery, ALFWorld, and WebShop, whereas policy updates matter relatively more on HotpotQA, with the contrast most pronounced on RealResearchQuery and ALFWorld. Notably, on HotpotQA, removing both causes a larger drop than the sum of the individual drops, showing that the two designs are not simply additive but reinforce each other by jointly aligning credit assignment and policy updates.

\subsection{In-Depth Analysis}
Unless otherwise specified, all in-depth analyses use Qwen3-4B-Instruct-2507 as the backbone model.

\paragraph{Training Dynamics.}
Figure~\ref{fig:training_curves} compares the training dynamics on HotpotQA and ALFWorld. CAPO achieves the highest rewards for most of training on both tasks, showing a consistent advantage throughout optimization. On HotpotQA, it generally exhibits lower critic loss than PPO after the initial updates. On ALFWorld, PPO's critic loss becomes highly unstable, whereas CAPO remains more controlled while attaining stronger rewards. Together, these results suggest that CAPO supports effective policy learning and stable optimization across different agent tasks.
On HotpotQA, CAPO reaches a reward above 0.5 by roughly 100 updates and remains at that level or higher, while PPO fluctuates around it. On ALFWorld, CAPO is the only method whose reward stays near 0.9 in the final training segment, whereas PPO remains below 0.6. The separation between the CAPO and PPO critic-loss curves persists in the latter half of training on both tasks, rather than arising from a single transient update.

\begin{figure*}[t]
\centering
\begin{minipage}[t]{0.48\textwidth}
\vspace{0pt}
\centering
\small
\captionof{table}{Effect of policy-ratio formulations (omitting $\theta$ for brevity) under action-level credit assignment. We compare token-wise ratios; product, mean, and square-root aggregations of token log-ratios within an action; and the action-aware centered square-root aggregation with batch-mean restoration. We report the seen win rate on ALFWorld and task score on WebShop.}
\label{tab:policy_ratio}
\resizebox{\linewidth}{!}{
\begin{tabular}{l|ccc}
\toprule
\textbf{Policy Update}
& \textbf{Log-Ratio Formulation}
& \textbf{ALFWorld}
& \textbf{WebShop} \\
\midrule
Token-wise
& $z_i^{(t)}$
& 84.29 & 73.80 \\

Product aggr.
& $\sum_{i=1}^{L_t} z_i^{(t)}$
& 74.29 & 68.72 \\

Mean aggr.
& $\frac{1}{L_t}\sum_{i=1}^{L_t} z_i^{(t)}$
& 89.29 & 75.14 \\

Sqrt. aggr.
& $\frac{1}{\sqrt{L_t}}\sum_{i=1}^{L_t} z_i^{(t)}$
& 90.71 & 76.18 \\

\textbf{Act.-aware}
& $\hat{\mu}
+\frac{1}{\sqrt{L_t}}
\sum_{i=1}^{L_t}
\left(z_i^{(t)}-\hat{\mu}\right)$
& \textbf{92.86} & \textbf{78.12} \\
\bottomrule
\end{tabular}
}
\end{minipage}\hfill
\begin{minipage}[t]{0.48\textwidth}
\vspace{0pt}
\centering
\includegraphics[width=\linewidth]{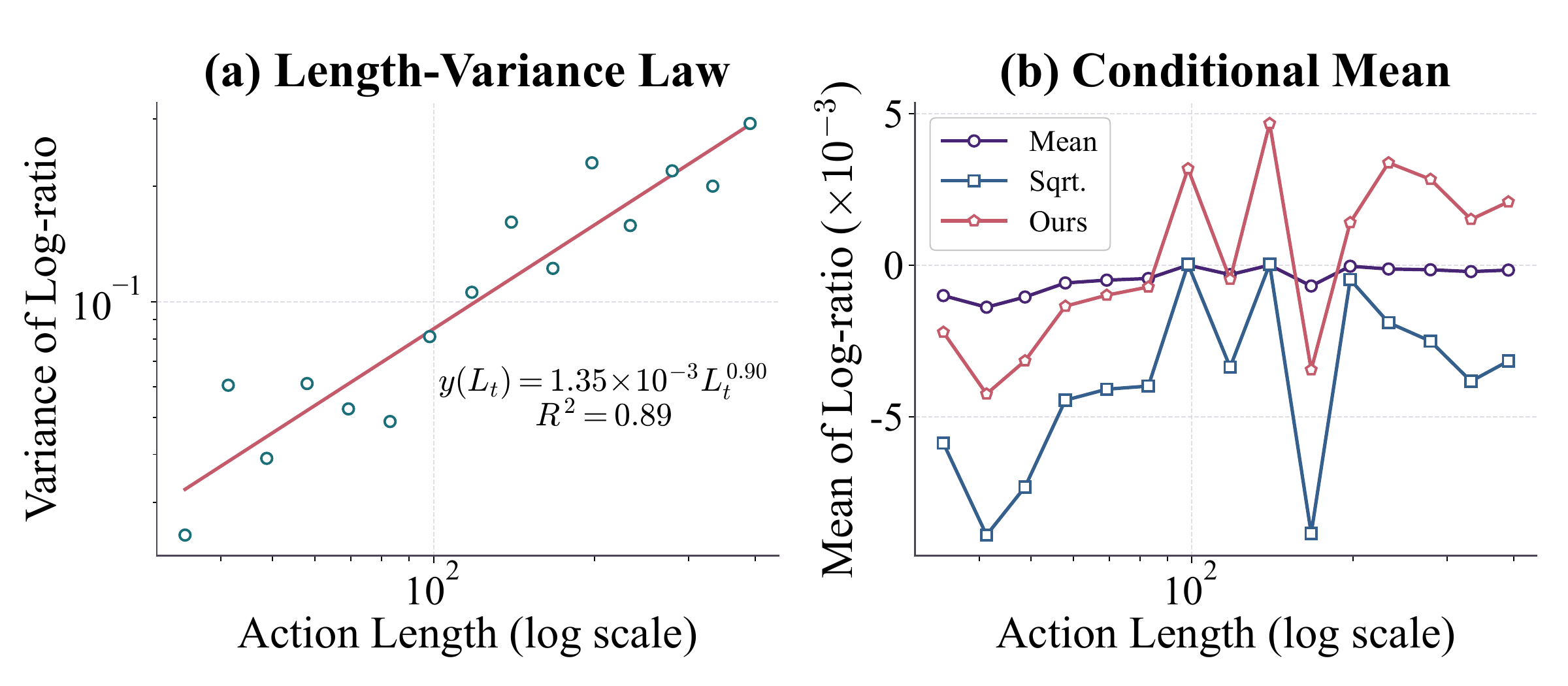}
\captionof{figure}{Policy ratio statistics across different action lengths during CAPO training on WebShop. We report the variance of $\sum_{i=1}^{L_t}\bigl(z_i^{(t)}(\theta)-\hat{\mu}(\theta)\bigr)$ and the mean log-ratio conditioned on $L_t$ under mean, sqrt., and action-aware aggregation.}
\label{fig:length_variance}
\end{minipage}
\end{figure*}

\begin{wraptable}{r}{0.52\linewidth}
\centering
\caption{Effect of GAE granularity across different $\gamma$ on WebShop. Act. Tokens denotes average action tokens per trajectory, and $\Delta$ Score is the relative change from $\gamma=1.00$.}
\label{tab:credit_assignment}
\small
\resizebox{\linewidth}{!}{
\begin{tabular}{c|cccc}
\toprule
\textbf{$\gamma$} & \textbf{Act. Tokens} & \textbf{Avg. Steps}
& \textbf{Score} & \textbf{$\Delta$ Score (\%)} \\
\midrule
\multicolumn{5}{l}{\textit{Token-level GAE}} \\
\hspace{0.5em}1.00 & 237.06 & 4.63 & 73.74 & -- \\
\hspace{0.5em}0.99 & 248.41 & 4.46 & 71.86 & -2.55 \\
\hspace{0.5em}0.95 & 275.92 & 4.25 & 63.02 & -14.54 \\
\midrule
\multicolumn{5}{l}{\textit{Act.-level GAE}} \\
\hspace{0.5em}1.00 & 140.08 & 4.79 & 74.83 & -- \\
\hspace{0.5em}0.99 & 143.37 & 4.94 & \textbf{78.12} & \textbf{+4.40} \\
\hspace{0.5em}0.95 & 183.61 & 4.18 & 73.46 & -1.83 \\
\bottomrule
\end{tabular}
}
\end{wraptable}

\paragraph{Credit Assignment.}
Table~\ref{tab:credit_assignment} studies GAE granularity on WebShop under different $\gamma$ values, with $\lambda$ fixed to 1.00. Action-level GAE consistently outperforms token-level GAE under matched settings, including $\gamma=1.00$, where credit is propagated without decay. Notably, token-level GAE at $\gamma=1.00$ performs similarly to action-level GAE at $\gamma=0.95$, despite the latter applying stronger discounting. Action-level GAE also remains more stable as $\gamma$ decreases, whereas token-level GAE is more sensitive because credit is propagated over much longer token sequences. Across settings, action-level GAE achieves better performance with fewer action tokens and comparable average interaction steps.
At $\gamma=0.99$, action-level GAE improves WebShop from 71.86 to 78.12 with shorter actions (143.37 vs. 248.41 tokens). It remains superior at $\gamma=0.95$ (73.46 vs. 63.02), indicating the gain is not due to longer trajectories.

\paragraph{Policy Updates.}
We examine how different policy ratio formulations affect policy updates under the action-level credit assignment. Table~\ref{tab:policy_ratio} shows that product aggregation underperforms even the token-wise ratio. Length-scaled aggregation substantially improves both tasks, with square-root aggregation consistently surpassing mean aggregation. The complete action-aware ratio further achieves the best results, validating both square-root calibration and batch-mean correction. Figure~\ref{fig:length_variance} provides supporting evidence from training statistics: the variance of centered log-ratio sums scales approximately as $L_t^{0.90}$, broadly supporting the square-root calibration intuition, while batch-mean restoration mitigates the length-dependent negative shift of direct square-root aggregation.
The action-aware ratio improves ALFWorld Seen from 90.71 to 92.86 and WebShop from 76.18 to 78.12 over square-root aggregation. Product aggregation is weaker (74.29 and 68.72), while mean aggregation only partially recovers. The mean plot shows that action-aware scaling remains length-stable, avoiding the downward shift of square-root aggregation.

\paragraph{Clipping Threshold Sensitivity.}
Table~\ref{tab:epsilon_sensitivity} evaluates CAPO under clipping thresholds on ALFWorld and WebShop. Unlike token-wise PPO, CAPO applies clipping to aggregated action-level policy ratios, whose scale is affected by action length, motivating a smaller clipping threshold. Across both tasks, increasing $\epsilon$ reduces the out-of-range and active clipping fractions as the clipping interval widens. Performance remains comparable from $3\times10^{-4}$ to $3\times10^{-3}$, indicating that the results are not tied to a threshold. Looser thresholds reduce clipping and degrade performance, suggesting that CAPO benefits from sufficiently tight constraints on policy updates.
The default $\epsilon=10^{-3}$ achieves the best performance (92.86 Seen, 78.12 WebShop). Larger $\epsilon$ values reduce active clipping and degrade scores, with $10^{-1}$ nearly eliminating clipping (0.6\%, 0.7\%) and dropping performance to 83.57 and 69.33, showing that overly loose clipping weakens optimization.

\begin{table}[t]
\centering
\small
\caption{Clipping threshold sensitivity of CAPO. OOR is the percentage of action-aware ratios outside $[1-\epsilon,1+\epsilon]$, while Active is the percentage that activates clipping in the policy objective. Seen and Score mean each task performance.}
\label{tab:epsilon_sensitivity}
\resizebox{0.9\linewidth}{!}{
\begin{tabular}{c|ccc|ccc}
\toprule
\multirow{2}{*}{$\boldsymbol{\epsilon}$}
& \multicolumn{3}{c|}{\textbf{ALFWorld}}
& \multicolumn{3}{c}{\textbf{WebShop}} \\
\cmidrule(lr){2-4}\cmidrule(lr){5-7}
& \textbf{OOR Fraction} & \textbf{Active Fraction} & \textbf{Seen}
& \textbf{OOR Fraction} & \textbf{Active Fraction} & \textbf{Score} \\
\midrule
$3\times10^{-4}$ & 66.8\% & 33.7\% & 91.43 & 70.5\% & 35.6\% & 76.84 \\
$1\times10^{-3}$ & 48.7\% & 24.5\% & \textbf{92.86} & 52.3\% & 26.4\% & \textbf{78.12} \\
$3\times10^{-3}$ & 29.4\% & 14.8\% & 92.14 & 32.8\% & 16.5\% & 77.46 \\
$1\times10^{-2}$ & 10.8\% & 5.4\%  & 89.29 & 12.6\% & 6.3\%  & 74.95 \\
$1\times10^{-1}$ & 1.1\%  & 0.6\%  & 83.57 & 1.4\%  & 0.7\%  & 69.33 \\
\bottomrule
\end{tabular}
}
\end{table}

\begin{wrapfigure}{r}{0.52\linewidth}
\centering
\includegraphics[width=\linewidth]{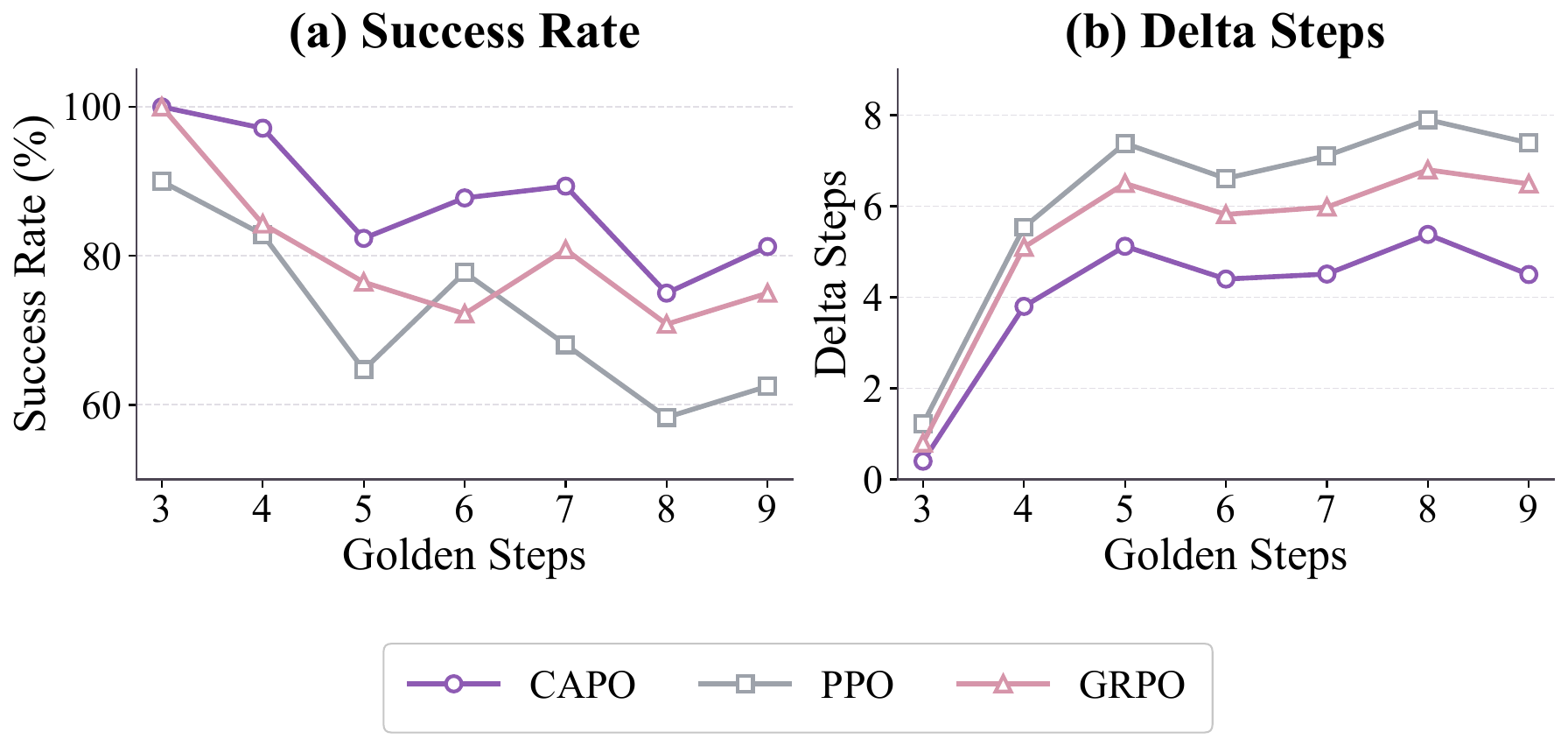}
\caption{ALFWorld performance by task horizon, grouped by human-annotated golden steps. Delta steps measure the gap between actual and golden step counts.
}
\label{fig:step_compare}
\end{wrapfigure}

\paragraph{Performance by Task Horizon.}
We group ALFWorld validation tasks by their human-annotated golden steps, defined as the minimum verified action sequence for task completion. Figure~\ref{fig:step_compare} reports the success rate and the gap between agent and golden steps. CAPO matches or exceeds PPO and GRPO across task horizons, with consistent gains on longer tasks. It also produces smaller step gaps, suggesting more efficient trajectories during task execution that remain closer to human-verified paths across long-horizon task groups.
Across every group from three to nine golden steps, CAPO has the lowest delta-step curve, including the longer groups where PPO and GRPO remain around six or more extra steps. Its success-rate curve is also never below either baseline: the advantage over PPO is particularly visible for five and seven-to-nine-step tasks.

\begin{wraptable}{r}{0.52\linewidth}
\centering
\small
\caption{Tool-use behavior on the academic paper search task with Qwen3-4B-Instruct-2507. We report the average number of action tokens per complete trajectory, search calls, expansion calls, and F1@all.}
\label{tab:tool_analysis}
\resizebox{\linewidth}{!}{
\begin{tabular}{l|cccc}
\toprule
\textbf{Method} & \textbf{Act. Tokens} & \textbf{Search} & \textbf{Expand} & \textbf{F1@all} \\
\midrule
PPO     & 2719.78 & 1.98 & 16.76 & 0.303 \\
GRPO    & 2951.57 & 2.74 & 15.48 & 0.294 \\
GiGPO   & 2814.79 & \textbf{4.57} & 16.54 & 0.306 \\
ST-PPO & 2798.10 & 3.85 & 17.17 & 0.302 \\
\textbf{CAPO}  & \textbf{2646.35} & 3.21 & \textbf{19.84} & \textbf{0.318} \\
\bottomrule
\end{tabular}
}
\end{wraptable}

\paragraph{Tool-Use Behavior.}
Table~\ref{tab:tool_analysis} compares retrieval-tool usage in academic paper search. CAPO obtains the highest F1@all while using the fewest action tokens and the most expansion calls among the compared methods. GiGPO issues more search calls, whereas CAPO allocates more interactions to citation and reference expansion. These statistics provide a behavioral view of the performance differences reported in the Experiments section.
CAPO reaches 0.318 F1@all, exceeding the next-best value of 0.306, while using 2646.35 action tokens per trajectory---fewer than every baseline. Its 19.84 expansion calls are also higher than the strongest baseline count of 17.17, whereas its 3.21 search calls are below GiGPO's 4.57. This pattern indicates that the improvement is associated with more use of citation/reference expansion, not with a larger overall action-token budget.

\begin{wraptable}{r}{0.52\linewidth}
\centering
\small
\caption{Training time per iteration. Units are seconds per iteration (s/iteration).}
\label{tab:appendix_training_computation}
\resizebox{\linewidth}{!}{
\begin{tabular}{lccc}
\toprule
\textbf{Method} & \textbf{Rollout Time} & \textbf{Actor Update} & \textbf{Critic Update} \\
\midrule
PPO & 14.12 & 76.65 & 104.60 \\
CAPO & 14.01 & 74.35 & 101.18 \\
\bottomrule
\end{tabular}
}
\end{wraptable}

\paragraph{Training Cost.}
CAPO introduces only lightweight operations beyond PPO. Its measured per-iteration time on the same server with 8 NVIDIA H100 GPUs described above is 189.54 seconds, compared with 195.37 seconds for PPO on WebShop. Action-level credit assignment uses value predictions only at state boundaries and runs GAE over interaction steps rather than generated tokens. Action-aware policy updates add only simple reductions and length calibration, while stop-gradient on $\hat{\mu}$ introduces no additional backpropagation or parameter updates. Unlike critic-free group-based methods, CAPO requires critic training, which we view as a trade-off for finer-grained credit assignment and improved performance.
Table~\ref{tab:appendix_training_computation} shows that the total difference is reflected in each measured phase: rollout takes 14.01 seconds for CAPO versus 14.12 for PPO, actor update takes 74.35 versus 76.65 seconds, and critic update takes 101.18 versus 104.60 seconds. Thus, the reported training-time comparison does not indicate a shifted cost from rollout into the update phases under this setting.

\begin{figure*}[t]
\centering
\includegraphics[width=\linewidth]{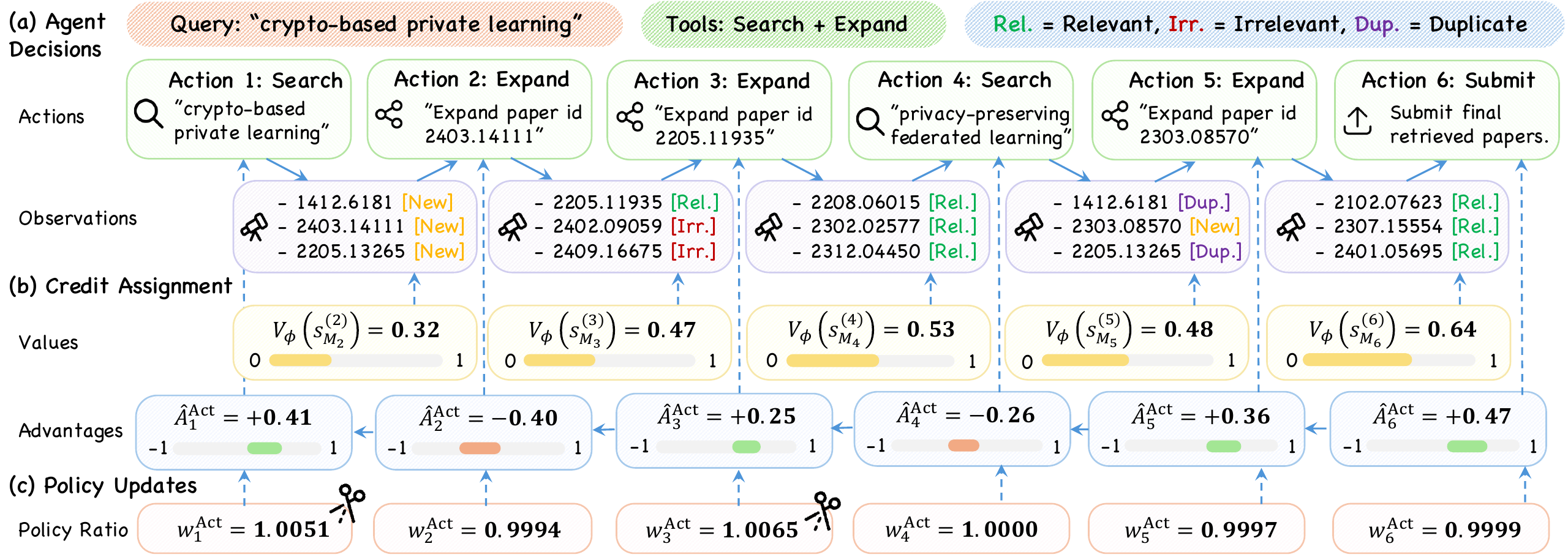}
\caption{
Case study on academic paper search. Action-level credit assignment propagates credit across interaction steps, while action-aware policy updates use a length-calibrated ratio to clip complete actions. Scissors mark ratios that activate clipping.
}
\label{fig:case}
\end{figure*}

\FloatBarrier

\paragraph{Case Study.}
Figure~\ref{fig:case} illustrates CAPO on a paper search trajectory. Values are estimated at state boundaries, and action-level GAE recursively propagates credit across the interaction sequence, assigning advantages to complete search and expansion actions. In the example, the agent alternates between search and citation/reference expansion actions, so each action receives a separate value difference and advantage rather than sharing a trajectory-level signal. Action-aware policy updates aggregate token-level changes within each action into a length-calibrated ratio for clipping, preventing the update scale from being dominated by action length. Together, these components align credit assignment and policy updates with agent decisions.

\section{Conclusion}

We propose CAPO, a critic-guided action-aligned policy optimization method for advancing LLM agent capabilities. CAPO mitigates the granularity mismatch between agent decisions and token-level optimization through two designs: action-level credit assignment through state value estimation at the action boundary and action-aware policy updates through a length-calibrated policy ratio. Together, these designs achieve fine-grained credit assignment and consistent policy updates while retaining token-level gradient computation. Across multi-hop QA, academic paper search, and text-world action tasks, CAPO consistently outperforms representative RL methods. Ablations and analyses validate both designs and reveal how action-aligned optimization improves multi-turn RL training. These results highlight complete actions as optimization units for training capable LLM agents.

\bibliographystyle{plainnat}
\bibliography{main}

\appendix

\section{Theoretical Analysis}
\label{app:theoretical_analysis}

\subsection{Detailed Step-Level MDP Transitions}

We model agent--environment interaction as a step-level MDP $\mathcal{M}_{\mathrm{Step}}=(\mathcal{S},\mathcal{A},\mathcal{P}_{\mathrm{Step}},\mathcal{R},\gamma)$. For brevity, let $s_t\triangleq s^{(t)}_{1:M_t}$ denote the state and $a_t\triangleq a^{(t)}_{1:L_t}$ the complete environment-facing action at step $t$. The action probability factorizes autoregressively as
\begin{equation}
\pi_\theta\!\left(a_t\mid s_t\right)
=
\prod_{i=1}^{L_t}
\pi_\theta\!\left(a_i^{(t)}\mid s_t,a_{<i}^{(t)}\right).
\end{equation}
Token decoding realizes $a_t$ internally but does not trigger environment transitions. After executing the complete action, the environment returns reward $r_t$ and observation $o_{t+1}$, from which the next state is constructed:
\begin{equation}
\begin{aligned}
(r_t,o_{t+1})
&\sim
\mathcal{P}_{\mathrm{env}}\!\left(\cdot\mid s_t,a_t\right),\\
s_{t+1}
&=
\mathcal{U}\!\left(s_t,a_t,o_{t+1}\right),
\end{aligned}
\end{equation}
where $\mathcal{U}$ denotes state construction. These operations induce $\mathcal{P}_{\mathrm{Step}}$ and $\mathcal{R}$.
Thus, the environment advances once per complete action rather than once per generated token.

\subsection{Action-Level Value and Advantage Estimation}

Because the environment transitions after a complete action, the corresponding action value is defined over the complete interaction step:
\begin{equation}
Q_t^{\mathrm{Act}}\!\left(s_t,a_t\right)
=
\mathbb{E}\!\left[
\sum_{k=t}^{T}\gamma^{k-t}r_k
\,\middle|\,
s_t,a_t
\right].
\end{equation}
The state value is the expected return before a particular action is specified, or equivalently the action value averaged over complete actions sampled from the current policy:
\begin{equation}
V_t^{\mathrm{Act}}\!\left(s_t\right)
=
\mathbb{E}_{a_t\sim\pi_\theta(\cdot\mid s_t)}
\!\left[
Q_t^{\mathrm{Act}}\!\left(s_t,a_t\right)
\right].
\end{equation}
Accordingly, the population action-level advantage compares the realized complete action with its pre-action expectation:
\begin{equation}
A_t^{\mathrm{Act}}
=
Q_t^{\mathrm{Act}}\!\left(s_t,a_t\right)
-
V_t^{\mathrm{Act}}\!\left(s_t\right).
\end{equation}

This definition places value estimation at the decision boundary before action generation. In a causal Transformer, the representation at the final state token is conditioned on the complete state prefix but not on the action. CAPO therefore parameterizes the state value as $V_\phi(s_t)\triangleq V_\phi(s^{(t)}_{M_t})$. A value prediction at an action-side position is already conditioned on at least a partial realization of the current action, while a prediction after the full action is closer to an action-conditioned value associated with $Q_t^{\mathrm{Act}}$. CAPO estimates the advantage using GAE along the interaction-step timeline:
\begin{equation}
\begin{aligned}
\hat A_t^{\mathrm{Act}}
&=
\sum_{l=0}^{T-t}(\gamma\lambda)^l\delta_{t+l},\\
\delta_t
&=
r_t+\gamma V_\phi(s_{t+1})
-V_\phi(s_t),
\end{aligned}
\end{equation}
with the value after a terminal step set to zero. The resulting $\hat A_t^{\mathrm{Act}}$ is assigned to the complete action and shared by its valid action tokens during policy optimization.

\subsection{Analysis of Action-Aligned Policy Optimization}

\paragraph{Policy-Ratio Construction for Action-Aware Updates.}
Let $z_i^{(t)}(\theta)=\log\pi_\theta(a_i^{(t)}\mid s_t,a_{<i}^{(t)})-\log\pi_{\mathrm{old}}(a_i^{(t)}\mid s_t,a_{<i}^{(t)})$ denote the token-level log-ratio. The product, geometric-mean, and direct square-root aggregations have the following log forms:
\begin{equation}
\begin{aligned}
\log w_t^{\mathrm{Prod}}
&=\sum_{i=1}^{L_t}z_i^{(t)},\\
\log w_t^{\mathrm{Mean}}
&=\frac{1}{L_t}\sum_{i=1}^{L_t}z_i^{(t)},\\
\log w_t^{\mathrm{Sqrt}}
&=\frac{1}{\sqrt{L_t}}\sum_{i=1}^{L_t}z_i^{(t)}.
\end{aligned}
\end{equation}
The product is the exact complete-action likelihood ratio, but its log-ratio accumulates over all action tokens and therefore becomes increasingly sensitive to action length. Geometric-mean aggregation controls this accumulation more strongly, but can concentrate long-action ratios near one and reduce their clipping frequency. Square-root aggregation lies between these two cases, retaining within-action accumulation while moderating its dependence on length.

CAPO additionally centers token log-ratios before square-root aggregation and restores their overall center:
\begin{equation}
\begin{aligned}
\log w_t^{\mathrm{Act}}(\theta)
&=
\hat\mu(\theta)
+\frac{1}{\sqrt{L_t}}
\sum_{i=1}^{L_t}\left(z_i^{(t)}(\theta)-\hat\mu(\theta)\right),\\
\hat\mu(\theta)
&=
\operatorname{sg}\!\left(
\frac{\sum_{b\in\mathcal B}\sum_{i=1}^{L_b}z_i^{(b)}(\theta)}
{\sum_{b\in\mathcal B}L_b}
\right).
\end{aligned}
\end{equation}

\paragraph{Role of Batch-Mean Centering.}
$\hat\mu(\theta)$ is a token-weighted empirical estimate of the average log-ratio between the current and rollout policies, rather than an additional learned parameter. Centering removes the shared policy shift before length scaling, while restoring $\hat\mu(\theta)$ retains the overall ratio center. At the rollout policy, all token log-ratios and $\hat\mu(\theta)$ are zero, so $w_t^{\mathrm{Act}}(\theta)=1$. As an empirical statistic, $\hat\mu(\theta)$ may vary with the composition of the current optimization set and is used as a practical calibration term rather than an exact population quantity. The stop-gradient operation allows it to affect the forward ratio and clipping decision without introducing derivative terms through the shared statistic. In particular, this prevents one step from receiving gradients contributed by other steps through the batch mean.

\paragraph{Gradient Interpretation and Relation to PPO.}
Treating $\hat A_t^{\mathrm{Act}}$ as fixed during the actor update, the gradient before clipping is activated is
\begin{equation}
\begin{aligned}
\nabla_\theta\!\left(
w_t^{\mathrm{Act}}(\theta)\hat A_t^{\mathrm{Act}}
\right)
&=
\frac{w_t^{\mathrm{Act}}(\theta)\hat A_t^{\mathrm{Act}}}{\sqrt{L_t}}
\\
&\quad\cdot
\sum_{i=1}^{L_t}
\nabla_\theta\log\pi_\theta\!\left(a_i^{(t)}\mid s_t,a_{<i}^{(t)}\right).
\end{aligned}
\end{equation}
Thus, the token score gradients within an action are combined under the same action advantage, action ratio, and clipping decision, with $L_t^{-1/2}$ calibrating the magnitude across action lengths. CAPO retains the clipped-surrogate structure of PPO but changes the update granularity: token-wise PPO weights and clips token ratios individually, whereas CAPO's action-aware policy updates construct one ratio and one clipping decision for the complete action. When $L_t=1$, the action-aware ratio reduces to the standard token ratio. Near the rollout policy, its per-action gradient follows the summed action score direction while applying a length-dependent rescaling, which intentionally prevents longer actions from dominating through token count.

Because $w_t^{\mathrm{Act}}(\theta)$ is a length-calibrated surrogate rather than the exact complete-action likelihood ratio, it should not be interpreted as an unbiased importance-sampling correction, and a TRPO-style guarantee does not directly apply. The analysis above instead clarifies how CAPO modifies the scale, gradient weighting, and clipping granularity of PPO to align policy updates with complete actions.

\section{Detailed Experimental Settings}
\label{app:experimental_settings}

\subsection{Datasets Preparation}

\paragraph{Multi-hop QA.}
The multi-hop QA setting uses HotpotQA \cite{yang2018hotpotqa} in the distractor setting as the in-domain benchmark. We train on the official HotpotQA training split with 90,447 questions and evaluate in-domain performance on the development split with 7,405 questions, where each question is paired with 10 Wikipedia context paragraphs. To evaluate cross-dataset generalization, we further test on the development splits of 2Wiki \cite{ho2020constructing} and MuSiQue \cite{trivedi2022musique}, containing 12,576 and 2,417 questions respectively; these examples are used only for evaluation and do not participate in parameter updates on HotpotQA. The reward combines the normalized exact-match score between the predicted answer and the gold answer with an intermediate formatting reward for valid \texttt{search(query)} calls; malformed tool-call outputs receive format feedback. We report answer accuracy (Acc).

\paragraph{Academic Paper Search.}
Academic paper search follows the multi-turn paper discovery setting studied in PaperScout \cite{pan2026paperscout}. The query data are built from RealResearchQuery \cite{he2025pasa}, with 33,551 research queries for training and 50 queries for testing. Retrieved papers are evaluated against query-level relevance annotations. Reward and inference scoring use pasa-7b-selector, a Qwen2.5-7B based model optimized for relevance assessment, which is not exposed as an agent tool. For each newly discovered paper, the selector takes the research query together with the paper title and abstract, and returns a relevance score. During training, papers with near-zero selector scores are discarded, and each valid \texttt{search(query)} or \texttt{expand(paper\_id)} call receives an intermediate formatting reward; each search or expansion step additionally receives the sum of the top three newly discovered relevance scores, minus optional action cost. Repeated search queries and malformed, invalid, or repeated expansions receive format feedback and a penalty of \(-0.5\). During inference, the same selector scores the final paper pool, and we report post-threshold F1@all and Recall@all against the annotated arXiv IDs.

\paragraph{ALFWorld.}
ALFWorld \cite{shridhar2020alfworld} is a text-based embodied household task benchmark. We filter the official household task data to solvable tasks from supported task types and use 3,553 training tasks, 140 valid-seen tasks, and 134 valid-unseen tasks, covering six task families: examining objects under light, pick-and-place, clean-and-place, cool-and-place, heat-and-place, and pick-two-and-place. Each episode provides a goal instruction and is evaluated by whether the agent completes the specified household task. The reward combines the terminal task-success signal with an intermediate formatting reward when the agent emits a valid admissible-command tool call; malformed outputs receive format feedback. We report win rate on the seen and unseen validation splits.

\paragraph{WebShop.}
WebShop \cite{yao2022webshop} is a text-based e-commerce navigation benchmark in which each goal specifies a shopping instruction. We use the full product setting, which contains approximately 1.18 million products and 12,087 predefined shopping goals. The split uses 11,587 goals for training and 500 goals for development. The reward combines the WebShop task score, computed from whether the purchased item satisfies the instruction and matches required product attributes, with an intermediate formatting reward for valid environment-action tool calls; purchase completion provides an additional success signal, and malformed outputs receive format feedback. We report both average task score (Score) and purchase success rate (Succ.).

\begin{table*}[t]
\centering
\small
\caption{Key optimization hyperparameters used in the main experiments. The critic learning rate applies to critic-based methods, including PPO, ST-PPO, and CAPO.}
\label{tab:appendix_training_hparams}
\resizebox{0.8\textwidth}{!}{
\begin{tabular}{lcccc}
\toprule
\textbf{Hyperparameter} & \textbf{HotpotQA} & \textbf{RealResearchQuery} & \textbf{ALFWorld} & \textbf{WebShop} \\
\midrule
Actor learning rate & \(1\times10^{-6}\) & \(1\times10^{-6}\) & \(1\times10^{-6}\) & \(1\times10^{-6}\) \\
Critic learning rate & \(1\times10^{-5}\) & \(1\times10^{-5}\) & \(1\times10^{-5}\) & \(1\times10^{-5}\) \\
Max prompt length & 10240 & 10240 & 8192 & 16384 \\
Max response length & 1024 & 4096 & 4096 & 4096 \\
Effective batch size & 128 & 128 & 128 & 128 \\
Group rollout number & 8 & 8 & 8 & 8 \\
Actor micro-batch size & 4 & 4 & 4 & 4 \\
Max environment steps & 5 & 5 & 20 & 15 \\
\bottomrule
\end{tabular}
}
\end{table*}

\subsection{RL Environment Construction}

\paragraph{Multi-hop QA.}
The multi-hop QA environment provides a dense retrieval interface over benchmark-specific Wikipedia paragraph indexes. For HotpotQA, the retrieval corpus is built by deduplicating all distractor context paragraphs from the training and development splits, yielding 509,308 passages. For cross-dataset evaluation, 2Wiki and MuSiQue use separate indexes built from the official full Wikipedia paragraph corpus with 5,902,082 passages and the deduplicated union of MuSiQue question-linked passages with 139,416 passages, respectively. We encode passages with \texttt{BAAI/bge-large-en-v1.5} and index them with FAISS. At each step, the agent observes the question, retrieved passages, previous search queries, and format feedback. It can issue up to 4 parallel \texttt{search(query)} calls per step, and the maximum horizon is 5 steps. When the agent has enough evidence, it terminates by outputting a short answer inside \texttt{<answer>...</answer>} tags.

\paragraph{Academic Paper Search.}
The academic paper search environment maintains a paper pool for each research query and exposes two tools: \texttt{search(query)}, which retrieves papers from an academic search service, and \texttt{expand(paper\_id)}, which expands the citations and references of a paper already in the pool. The paper corpus is constructed from a January 2026 Semantic Scholar snapshot by retaining arXiv papers with abstracts, resulting in approximately 3 million papers and 30 million in-corpus citation edges. The retrieval API supports sparse retrieval with a SQLite FTS5 full-text index using BM25 ranking, dense retrieval with Qdrant and BGE-M3 embeddings, and hybrid retrieval through reciprocal-rank fusion. The default search returns 10 papers; citation expansion returns up to 30 citing papers, and reference expansion returns up to 99 referenced papers. At each step, the agent observes the user query, the current paper pool, and previous search or expansion actions. It can make up to 5 parallel tool calls per step, and the maximum horizon is 5 steps.

\paragraph{ALFWorld.}
The ALFWorld environment is built on TextWorld. Each episode corresponds to a packaged TextWorld game file, and the runtime wrapper loads one game as an independent interaction instance. At each step, the wrapper returns the current textual observation and the admissible-command list, and the agent must submit exactly one command that matches the list through the environment-step tool. These commands include navigation and household operations such as moving, opening or closing objects, taking objects, and placing objects. Task success is determined by the \texttt{won} signal returned by the environment. For each agent step, the prompt is reconstructed from the current observation and history actions rather than accumulated as a full multi-turn dialogue, matching the step-level MDP formulation. We set the maximum horizon to 20 interaction steps.

\paragraph{WebShop.}
The WebShop environment is implemented as a self-hosted HTTP shopping simulator. It builds a SQLite product store and a Lucene full-text search index over the full product catalog. During training, the client interacts with the server through reset and step calls: reset initializes a shopping goal, and each step applies one executable action to the current page state. Available actions are generated dynamically from the current page, including keyword search, product clicks, option selection, back navigation, and purchase. The agent issues one action through the environment-step tool and receives the next page observation. Each agent step reconstructs the prompt from the current observation and recent action history rather than carrying a full dialogue. We set the maximum horizon to 15 interaction steps.

\subsection{Baselines}

\paragraph{ReAct-style Prompting.}
The prompting baseline evaluates the LLM without RL fine-tuning. It uses the same task prompts and tool schemas as the RL methods, following a ReAct-style format \cite{yao2022react} that interleaves reasoning and environment-facing actions. We set the evaluation temperature to 0.0. This baseline measures the policy's capability under the interaction protocol before optimization.

\paragraph{PPO.}
PPO \cite{schulman2017ppo} is implemented as the token-level GAE baseline. It uses a learned critic and estimates advantages at token granularity. This baseline keeps the conventional LLM RL view that generated tokens are the optimization unit, even though the environment state changes only after a complete interaction response.

\paragraph{Reinforce++.}
Reinforce++ \cite{hu2025reinforce++} is a critic-free token-level return baseline. It computes discounted returns over valid generated tokens, uses the same-prompt rollout-group mean as a baseline, and applies masked whitening.

\paragraph{GRPO.}
GRPO \cite{shao2024deepseekmath} is a critic-free group-relative baseline. For each prompt, it samples multiple rollouts and computes a trajectory-level relative advantage from the group rewards. We use 8 rollouts per prompt by default and reduce the effective prompt batch size accordingly to keep the update budget comparable.

\paragraph{RLOO.}
RLOO \cite{ahmadian2024back} is another trajectory-level group baseline. It computes each rollout's baseline from the average reward of the other rollouts in the same prompt group, leaving out the current rollout's own reward. The resulting advantage is still assigned at trajectory granularity.

\paragraph{GSPO.}
GSPO \cite{zheng2025group} is used as a sequence-level policy-loss baseline. In our implementation, it keeps the GRPO trajectory-level advantage estimator and replaces the token-level policy ratio with a sequence-level ratio objective. This isolates the effect of the sequence-level policy loss from CAPO's action-level credit assignment.

\paragraph{GiGPO.}
GiGPO \cite{feng2026gigpo} is a critic-free baseline that combines trajectory-level group comparison with grouped step-level information. It introduces a step-advantage component grouped by anchor observations, but does not learn a value function. We include it as a strong comparison for finer-grained agent credit assignment.

\paragraph{ST-PPO.}
ST-PPO \cite{li2025st} is a critic-based multi-turn PPO baseline. Following its original formulation, it replaces token-level importance sampling with a geometric-mean importance weight over each turn and further normalizes the resulting gradient using a clipping-bias correction. We map each turn to an interaction step and categorize its turn-level ratio as sequence aggregation in our comparison.

\subsection{Training Settings}

Our implementation uses Agent-R1 \cite{cheng2025agentrone} as the training framework, with veRL \cite{sheng2025hybridflow} as the backend RL engine and vLLM generation \cite{kwon2023efficient}. Experiments are run on a server with 8 NVIDIA H100 GPUs. Unless otherwise specified, we set the actor learning rate to \(1\times10^{-6}\), the critic learning rate to \(1\times10^{-5}\) for critic-based methods, the training batch size to $128$, and the actor micro-batch size to $4$ per GPU. Group-based baselines use $8$ rollouts per prompt and a batch size of $16$, yielding an effective batch size of $128$ for fair comparison. For policy clipping, we set $\epsilon=0.001$ for CAPO and $\epsilon=0.0003$ for GSPO; the remaining baselines follow the clipping configurations in their original papers. We use neither learning-rate warm-up nor a separate critic warm-up stage. For CAPO, we use discount factor $\gamma = 0.99$, GAE trace parameter $\lambda = 1.0$, and actor-side KL regularization coefficient \(0.001\). Within each benchmark, all RL methods share the same backbone model, training data, task description, tool interface, hyperparameter budget, and evaluation protocol; they differ only in their RL algorithm design. CAPO, PPO, and ST-PPO use a learned critic, while GRPO, RLOO, Reinforce++, GSPO, and GiGPO are critic-free baselines according to their definitions.

\section{Prompt Templates}
\label{app:prompt_templates}

This appendix lists the prompt templates used to instantiate the tool-calling interface described in the Experimental Setup subsection of the main paper's Experiments section. The \texttt{prompt} field stored in each dataset row contains the task input, such as the HotpotQA question, paper-search query, ALFWorld goal, or WebShop shopping instruction. During rollout, the agent flow reconstructs a step prompt from the current observation or retrieved evidence, action history, available tools or admissible actions, and expected output format. This construction matches the step-level MDP formulation: each prompt represents the interaction state, and each response is parsed as one environment-facing action. We preserve placeholders \texttt{\{observation\}} and \texttt{\{history\_actions\}} because they are populated at rollout time.

\section{Representative Trajectory}
\label{app:paper_search_trajectory}

To complement the case study in the main paper's In-Depth Analysis subsection, we include a shortened academic paper search trajectory. The example illustrates how the agent grows a paper pool through search and citation/reference expansion, while selecting complete actions at each interaction step. It is lightly edited from a real rollout: long observations, full paper lists, and repetitive parallel tool calls are abbreviated with ellipses, while the action-level decision pattern is preserved.

\section{Reproducibility Details}

All RL results are averaged over random seeds 41, 42, and 43. ReAct is evaluated once with temperature 0.0 using deterministic decoding. The hyperparameter comparisons cover $\gamma\in\{0.95,0.99,1.00\}$ and the policy-ratio variants reported in the policy ratio analyses. Experiments run on a server with eight NVIDIA H100 80GB HBM3 GPUs and dual Intel Xeon Platinum 8468 CPUs under Ubuntu 22.04.5 LTS. The software stack uses Python 3.10.20, CUDA 12.8, PyTorch 2.8.0+cu128, Transformers 4.57.6, veRL 0.7.0, and vLLM 0.11.0. The complete preprocessing, training and evaluation code is available in the supplementary material.

\section{Limitations}
Our work has several limitations. First, although CAPO has comparable per-iteration training time to PPO in our measured setting, it still requires critic inference and updates, unlike critic-free algorithms such as GRPO. Extending action-aligned optimization to critic-free RL algorithms is an important direction for future work. Second, the current study mainly focuses on single ReAct-style agent settings and does not explore more complex multi-agent or hierarchical systems. Whether action-aligned optimization remains effective in these settings requires further investigation. Third, we have not validated CAPO on very large backbone models, leaving its scalability to stronger LLM agents for future study. Finally, we only validate CAPO on interactive agent tasks, including multi-hop QA, academic paper search, and text-world action tasks. Its effectiveness for broader post-training scenarios, such as alignment and reasoning tasks, remains to be explored.

\begin{StrategyBox}[framegreen]{bggreen}{HotpotQA Prompt}
\refstepcounter{idx}
\label{pro:hotpotqa_prompt}

\begin{verbatim}
You are a research agent. Your goal is to answer the User Query using 
Wikipedia search evidence.
\end{verbatim}

\noindent\textbf{User Query}
\begin{verbatim}
{user_query}
\end{verbatim}
\noindent\textbf{History Actions}
\begin{verbatim}
{history_actions}
\end{verbatim}
\noindent\textbf{Retrieved Passages}
\begin{verbatim}
{passage_list}
\end{verbatim}
\noindent\textbf{Recent tool / format issues}
\begin{verbatim}
{tool_feedback}
\end{verbatim}
\noindent\textbf{Instructions}
\begin{verbatim}
- Analyze the Retrieved Passages and History Actions to determine the next set
of actions. Enclose your analysis of the state and decision logic within
`<analysis>...</analysis>` tags.
- You support parallel tool calling. You should output multiple tool calls in a
single step if several independent actions are valuable at the current state.
- Attend to the history actions and avoid repeating the same search queries.
When you can answer the question from the current passages, put the short final
answer inside `<answer></answer>` tags  instead of further tool calls.
\end{verbatim}

\noindent\textbf{Tool Definition}
\begin{verbatim}
{
  "type": "function",
  "function": {
    "name": "search",
    "description": "Search Wikipedia for passages relevant to the user question. Use
       natural-language or keyword queries; must differ from prior history queries
       when possible.",
    "parameters": {
      "type": "object",
      "properties": {
        "query": {
          "type": "string",
          "description": "A single search query (natural language or keywords). Must
             differ from all history queries when seeking new evidence."
        }
      },
      "required": ["query"]
    }
  }
}
\end{verbatim}
\end{StrategyBox}
\onecolumn

\begin{BreakableStrategyBox}[frameorange]{bgorange}{Paper Search Prompt}
\refstepcounter{idx}
\label{pro:paper_search_prompt}

\begin{verbatim}
You are a research agent. Your goal is to find papers relevant to the User Query.
\end{verbatim}

\noindent\textbf{User Query}
\begin{verbatim}
{user_query}
\end{verbatim}
\noindent\textbf{History Actions}
\begin{verbatim}
{history_actions}
\end{verbatim}
\noindent\textbf{Paper List}
\begin{verbatim}
{paper_list}
\end{verbatim}
\noindent\textbf{Instructions}
\begin{verbatim}
- Analyze the Paper List and History Actions to determine the next set of
actions. Enclose your analysis of the state and decision logic within
`<analysis>...</analysis>` tags.
- You support parallel tool calling. You should output multiple tool calls in a
single step if several independent actions are valuable at the current state.
- Attend to the history actions and avoid repeating the same search query or
expanding the same paper.
\end{verbatim}

\noindent\textbf{Tool Definitions}
\begin{verbatim}
[
  {
    "type": "function",
    "function": {
      "name": "search",
      "description": "Search for relevant papers with the hybrid retrieval API.",
      "parameters": {
        "type": "object",
        "properties": {
          "query": {
            "type": "string",
            "description": "A single search query in natural language or keywords.
              Must differ from all history queries."
          }
        },
        "required": ["query"]
      }
    }
  },
\end{verbatim}
\begin{verbatim}
  {
    "type": "function",
    "function": {
      "name": "expand",
      "description": "Expand from an existing paper by merging its citations and
        references to surface more related works.",
      "parameters": {
        "type": "object",
        "properties": {
          "paper_id": {
            "type": "string",
            "description": "The paper identifier of a paper already present in the
              current paper list."
          }
        },
        "required": ["paper_id"]
      }
    }
  }
]
\end{verbatim}
\end{BreakableStrategyBox}

\begin{BreakableStrategyBox}[frameorange]{bgorange}{Paper Search Selector Prompt}
\refstepcounter{idx}
\label{pro:paper_search_selector_prompt}

\noindent
\begin{verbatim}
You are an elite researcher in the field of AI, conducting research on {user_query}.
Evaluate whether the following paper fully satisfies the detailed requirements of
the user query and provide your reasoning. Ensure that your decision and reasoning
are consistent.

Searched Paper:
Title: {title}
Abstract: {abstract}

User Query: {user_query}

Output format: Decision: True/False
Reason:...
Decision:
\end{verbatim}
\end{BreakableStrategyBox}

\begin{StrategyBox}[frameblue]{bgblue}{ALFWorld Prompt}
\refstepcounter{idx}
\label{pro:alfworld_prompt}

\begin{verbatim}
You are acting in ALFWorld TextWorld. Choose exactly one command from the provided  
admissible commands each turn. You may reason briefly inside <think></think> before 
acting. Call the env\_step tool with a exact command. Do not output a final answer.
\end{verbatim}

\noindent\textbf{Current Observation}
\begin{verbatim}
{observation}
\end{verbatim}
\noindent\textbf{History Actions}
\begin{verbatim}
{history_actions}
\end{verbatim}
\noindent\textbf{Admissible Commands}
\begin{verbatim}
{admissible_commands}
\end{verbatim}
\noindent\textbf{Instructions}
\begin{verbatim}
- Think briefly about the current state inside `<think>...</think>` tags.
- Use exactly one command through the `env_step` tool.
- The command must exactly match one item from `Admissible Commands`.
- Follow ALFWorld TextWorld command style such as `go to dresser 1`,
  `take mug 1 from cabinet 3`, `use desklamp 1`.
- Use the official observation text as the source of truth.
- Do not output explanations or a final natural-language answer.
\end{verbatim}
\noindent\textbf{Output Format}
\begin{verbatim}
<think>
[Your brief reasoning about the current state and next command.]
</think>
<tool_call>
{{"name": "env_step", "arguments": {{"command": "[one admissible command]"}}}}
</tool_call>
\end{verbatim}

\noindent\textbf{Tool Definition}
\begin{verbatim}
{
  "type": "function",
  "function": {
    "name": "env_step",
    "description": "Execute one ALFWorld TextWorld command and return the next
       official observation.",
    "parameters": {
      "type": "object",
      "properties": {
        "command": {
          "type": "string",
          "description": "A single ALFWorld TextWorld command such as `go to dresser
              1`, `open cabinet 3`, `take mug 1 from cabinet 3`, `use desklamp 1`. It
            must exactly match one currently admissible command."
        }
      },
      "required": ["command"]
    }
  }
}
\end{verbatim}
\end{StrategyBox}

\begin{StrategyBox}[frameblue]{bgblue}{WebShop Prompt}
\refstepcounter{idx}
\label{pro:webshop_prompt}

\begin{verbatim}
You are acting in the WebShop text environment. Your goal is to find and buy
the product that best satisfies the shopping instruction. Use exactly one executable 
WebShop action each turn through the env\_step tool. Do not explain.
\end{verbatim}

\noindent\textbf{Shopping Instruction}
\begin{verbatim}
{instruction}
\end{verbatim}
\noindent\textbf{Current Observation}
\begin{verbatim}
{observation}
\end{verbatim}
\noindent\textbf{Recent History}
\begin{verbatim}
{recent_history}
\end{verbatim}
\noindent\textbf{Available Actions}
\begin{verbatim}
{available_actions}
\end{verbatim}
\noindent\textbf{Instructions}
\begin{verbatim}
- Use exactly one action through the `env_step` tool.
- The `command` must be one available action exactly, except replace `<your query>`
   in `search[<your query>]` with concise product keywords.
- Click product ASINs, option values, `Description`, `Features`, `Reviews`, `Back
    to Search`, `Back to Item`, or `Buy Now` only when listed above.
- Buy only when the selected product and options satisfy the shopping instruction.
\end{verbatim}

\noindent\textbf{Tool Definition}
\begin{verbatim}
{
  "type": "function",
  "function": {
    "name": "env_step",
    "description": "Execute one WebShop action and return the next  observation.",
    "parameters": {
      "type": "object",
      "properties": {
        "command": {
          "type": "string",
          "description": "A single action such as `search[wireless headphones]` or
          `click[Buy Now]`."
        }
      },
      "required": ["command"]
    }
  }
}
\end{verbatim}
\end{StrategyBox}

\begin{BreakableStrategyBox}[frameorange]{bgorange}{Paper Search Trajectory}
\refstepcounter{idx}
\label{pro:paper_search_trajectory}

\begin{verbatim}
User Query:
crypto-based private learning

==================== Step 1 ====================
<analysis>
The paper pool is empty. The query concerns privacy-preserving learning with
cryptographic techniques, so the first step should build a broad seed pool. Use
parallel searches that cover homomorphic encryption, federated learning,
secure computation, and differential privacy. No previous queries or expanded
papers need to be avoided yet.
</analysis>
<tool_call>
{"name": "search", "arguments": {"query": "crypto-based private learning"}}
</tool_call>
<tool_call>
{"name": "search", "arguments": {"query": "homomorphic encryption ML"}}
</tool_call>
...

==================== Step 2 ====================
Observation summary:
The pool now contains seed papers on encrypted inference, private training,
federated learning, and privacy-preserving neural networks.

<analysis>
Several retrieved papers directly match the query, especially works involving
homomorphic encryption and encrypted neural-network inference. To turn the seed
pool into a connected research graph, expand the most relevant new papers. Avoid
repeating the four search queries from Step 1.
</analysis>
<tool_call>
{"name": "expand", "arguments": {"paper_id": "1412.6181"}}
</tool_call>
<tool_call>
{"name": "expand", "arguments": {"paper_id": "2403.14111"}}
</tool_call>
...

==================== Step 3 ====================
Observation summary:
Expansion surfaces papers on FHE-based inference, encrypted training, transfer
learning, transformers, and activation-function approximations.

<analysis>
The pool is now richer but still concentrated around encrypted inference. Expand
new papers that diversify the graph toward training, transformer models, secure
pipelines, and systems. Skip papers already expanded in previous steps.
</analysis>
<tool_call>
{"name": "expand", "arguments": {"paper_id": "2205.11935"}}
</tool_call>
<tool_call>
{"name": "expand", "arguments": {"paper_id": "2402.09059"}}
</tool_call>
...

==================== Step 4 ====================
Observation summary:
New papers include CryptoDL, THE-X, MedBlindTuner, CURE, Pencil, SPEED, SoK
surveys, and secure split-learning or hybrid-encryption methods.

<analysis>
The graph now covers both foundational and recent work. The next useful action
is not another broad search, but targeted citation/reference expansion from
new high-signal papers. Prioritize papers that connect FHE with training,
fine-tuning, transformers, secret sharing, or split learning. Avoid expanding
papers already marked as explored.
</analysis>
<tool_call>
{"name": "expand", "arguments": {"paper_id": "1711.05189"}}
</tool_call>
<tool_call>
{"name": "expand", "arguments": {"paper_id": "2107.14338"}}
</tool_call>
...

==================== Step 5 ====================
Observation summary:
The pool contains a broad set of relevant candidates: FHE-based neural-network
inference, private training, graph neural networks, RNN training, hybrid
HE/secret-sharing systems, functional encryption for federated learning, and
survey papers.

<analysis>
At this point, most broad directions have been covered. Expand only the remaining
new papers that add missing subtopics, such as graph learning, recurrent models,
hybrid cryptographic protocols, or comparative surveys. This final step improves
recall while avoiding repeated searches and already-expanded papers.
</analysis>
<tool_call>
{"name": "expand", "arguments": {"paper_id": "2012.09364"}}
</tool_call>
<tool_call>
{"name": "expand", "arguments": {"paper_id": "2209.11904"}}
</tool_call>
...
\end{verbatim}
\end{BreakableStrategyBox}

\end{document}